\documentclass[journal]{IEEEtran}
\usepackage{amsmath,amssymb,amsfonts}
\usepackage{algorithmic}
\usepackage{algorithm}
\usepackage{array}
\usepackage[caption=false,font=normalsize,labelfont=sf,textfont=sf]{subfig}
\usepackage{textcomp}
\usepackage{stfloats}
\usepackage{url}
\usepackage{verbatim}
\usepackage{graphicx}
\usepackage{cite}
\usepackage{hyperref} 
\usepackage{multirow}
\usepackage{textcomp}
\usepackage{xcolor}
\usepackage{tabularx}
\usepackage{booktabs}
\usepackage{rotating} 
\usepackage{caption}

\usepackage{amsthm}
\usepackage{ragged2e} 
\usepackage{enumitem}
\newlist{tabitemize}{itemize}{1} 
\setlist[tabitemize]{label=\textbullet, nosep, leftmargin=*, 
                     before={\begin{minipage}[t]{\linewidth}\RaggedRight},
						after={\end{minipage}}}
\newtheorem{definition}{Definition}

\begin{document}

\title{Machine Unlearning: Solutions and Challenges}

\author{Jie Xu, Zihan Wu, Cong Wang~\IEEEmembership{Fellow,~IEEE,} and Xiaohua Jia~\IEEEmembership{Fellow,~IEEE}
\thanks{© 2024 IEEE.  Personal use of this material is permitted.  Permission from IEEE must be obtained for all other uses, in any current or future media, including reprinting/republishing this material for advertising or promotional purposes, creating new collective works, for resale or redistribution to servers or lists, or reuse of any copyrighted component of this work in other works.}
}

\markboth{Journal of \LaTeX\ Class Files,~Vol.~14, No.~8, August~2021}%
{Shell \MakeLowercase{\textit{et al.}}: A Sample Article Using IEEEtran.cls for IEEE Journals}


\maketitle

\begin{abstract}
	Machine learning models may inadvertently memorize sensitive, unauthorized, or malicious data, posing risks of privacy breaches, security vulnerabilities, and performance degradation. To address these issues, machine unlearning has emerged as a critical technique to selectively remove specific training data points' influence on trained models. This paper provides a comprehensive taxonomy and analysis of the solutions in machine unlearning. We categorize existing solutions into exact unlearning approaches that remove data influence thoroughly and approximate unlearning approaches that efficiently minimize data influence. By comprehensively reviewing solutions, we identify and discuss their strengths and limitations. Furthermore, we propose future directions to advance machine unlearning and establish it as an essential capability for trustworthy and adaptive machine learning models. This paper provides researchers with a roadmap of open problems, encouraging impactful contributions to address real-world needs for selective data removal. 
\end{abstract}


\section{Introduction}
\IEEEPARstart{T}{he} rapid expansion of Machine Learning (ML) has led to remarkable advancements in tasks such as image recognition~\cite{zhao2020exploring}, natural language processing~\cite{hirschberg2015advances}, and recommendation systems~\cite{wu2022graph}, significantly impacting various aspects of our lives~\cite{jordan2015machine}. However, the wide adoption of machine learning has raised significant concerns about potential privacy risks, security vulnerabilities, and accuracy degradation in dynamic settings~\cite{fu2021knowledge,tarun2023fast,jegorova2022survey}.  

To address these issues, machine unlearning has emerged as a promising technique. Machine unlearning refers to the process of selectively removing specific training data points and their influence on an already trained model, making the updated model behave the same as a model that was never trained on that data~\cite{xu2023machine}. It provides a `subtractive' capability to adapt models by removing unauthorized, malicious, or outdated data points without full retraining.

Machine unlearning plays an important role in ML models.
First, machine unlearning enforces privacy regulations and protects user privacy. Laws such as the European Union (EU), \textit{General Data Protection Regulation} (GDPR)~\cite{Voigt2017} and \textit{California Consumer Privacy Act} (CCPA)~\cite{CCPA2023} have introduced the `Right to be Forgotten,' allowing users to request removing their personal data from trained models~\cite{wang2019measure,chang2021privacy}. By enabling selective data removal, machine unlearning provides an efficient way to achieve personal data relevant to protect users' legal rights.

Second, machine unlearning enhances model security and robustness against adversarial attacks. Machine learning models are vulnerable to adversarial attacks such as data poisoning attacks, where adversaries inject crafted malicious data into the training dataset to manipulate the model's behavior~\cite{biggio2012poisoning, steinhardt2017certified}. By removing harmful manipulated data points, machine unlearning helps defend models against such vulnerabilities.

Third, machine unlearning improves the adaptability of models over time in dynamic environments. Models trained on static historical data can become outdated as the data distributions shift over time~\cite{gama2014survey}. For example, customer preferences may change in the recommendation system. By selectively removing outdated or unrepresentative data, machine unlearning enables the model to maintain performance even as the environment evolves~\cite{cao2015towards}.

Based on the residual influence of the removed data point, machine unlearning solutions can be categorized into \textit{Exact Unlearning} and \textit{Approximate Unlearning}~\cite{thudi2022necessity}. 
Exact unlearning aims to completely remove the influence of targeted data points from the model through algorithmic-level retraining ~\cite{thudi2022necessity, Yan2022ARCANE}. The advantage of this method is the model behaves as if the unlearned data had never been used. While providing strong guarantees of removal, exact unlearning usually demands extensive computational and storage resources and is primarily suitable for simpler models. 
On the other hand, approximate unlearning focuses on reducing the influence of targeted data points through efficient model parameter update~\cite{Yan2022ARCANE,thudi2022necessity}. While not removing influence thoroughly, approximate unlearning significantly reduces computational, storage, and time costs. It enables efficient unlearning of large-scale and complex models where exact unlearning is impractical.

In this paper, we provide a comprehensive overview of the solutions proposed for machine unlearning, covering pioneering and state-of-the-art techniques for both exact and approximate unlearning. We critically analyze existing solutions, highlight their advantages and limitations, identify research gaps, and suggest future directions. Our goal is to provide a useful roadmap that helps guide future work in developing adaptive and trustworthy ML systems, directly addressing evolving real-world challenges.

The main contributions of this paper are:
\begin{itemize}
	\item We provide a comprehensive taxonomy and structured overview of machine unlearning solutions, categorizing techniques into exact and approximate unlearning approaches. The taxonomy covers a broad range of existing works, establishing a structured understanding of this emerging field for researchers.
	\item We give an in-depth critical analysis of machine unlearning solutions, highlighting their strengths, limitations, and challenges. This analysis provides valuable insights into theoretical and practical obstacles, guiding future research toward impactful open problems.
	\item We identify critical issues in machine unlearning and suggest promising directions for future research. These suggestions expand the applicability of machine unlearning and address its limitations effectively.
\end{itemize}

The remainder of this paper is organized as follows. Section \ref{secback} provides background and preliminaries. Section \ref{secexact} covers exact unlearning, and Section \ref{secapp} explores approximate unlearning. Section \ref{secdis} offers a comprehensive discussion on critical issues of machine unlearning. Section \ref{secpotential} outlines potential future research directions. Finally, Section \ref{seccon} concludes by summarizing key findings.

\section{Background and Preliminaries}\label{secback}
\subsection{Machine Learning}
Machine learning develops algorithms that allow computers to automatically learn and improve from experience based on data~\cite{jordan2015machine}.  

\subsubsection{Supervised Learning}
Supervised learning is a core subset of machine learning, which is also a central focus of the current machine unlearning research. It involves training algorithms on a labeled dataset, $\mathcal{D} = \{(\mathbf{x}_i, y_i)\}_{i=1}^n$, where each example is a pair consisting of an input feature vector $\mathbf{x}_i$ and its corresponding output label $y_i$. 
Here, $\mathbf{x}_i \in \mathcal{X}$ denotes the input feature vector from the input space $\mathcal{X} \subset \mathbb{R}^d$, where $d$ is the dimension of the input features. $y_i \in \mathcal{Y}$ is the corresponding output variable from the output space $\mathcal{Y}$. The goal is to learn a model  $\mathcal{M}$ parameterized by $\mathbf{w}$ that maps input $\mathbf{x}$ to outputs $y$. This is achieved by minimizing a loss function $F(\mathcal{D}; \mathbf{w}) = \sum_{z\in\mathcal{D}}f(z; \mathbf{w})$ that quantifies the discrepancy between the predicted outputs $\hat{y}_i=\mathcal{M}(\mathbf{x}_i; \mathbf{w})$ and the true outputs $y_i$ over the training data. The optimal parameters $\mathbf{w}^*$ are obtained by:
\begin{equation}
	\mathbf{w}^*=\operatorname{argmin}\limits_{\mathbf{w}\in \mathcal{H}} F(\mathcal{D}; \mathbf{w})
\end{equation}
where $\mathcal{H}$ denotes the hypothesis space containing candidate models.

\subsubsection{Explainable Artificial Intelligence (AI) }
Explainable AI techniques aim to increase the transparency and explainability of complex models, making the relationship between training data and model predictions clear~\cite{doshi2017towards,arrieta2020explainable}. This facilitates us to understand how removing specific training data affects model predictions~\cite{Koh2017Understanding}. 

Influence function is a tool in explainable AI that can quantify individual training points' influence on a model~\cite{hampel1974influence}. The influence of a point $z'\in\mathcal{D}$ on model parameters $\mathbf{w}$ can be assessed by slightly increasing its weight during training:
\begin{equation}
	\hat{\mathbf{w}}_{\epsilon; z'} \stackrel{\text { def }}{=} \arg \min _{\mathbf{w} \in \mathcal{H}} \frac{1}{n} \sum_{i=1}^n f\left(z_i; \mathbf{w}\right)+\epsilon f(z'; \mathbf{w}).
\end{equation}

The influence function, represented by $\mathcal{I}$, calculates how changes in the weight of the data point $z'$ affect the model's parameters and is shown as Eq.(\ref{eqinf}).
\begin{equation} \label{eqinf}
	\mathcal{I}(z'; f, \hat{\mathbf{w}},\mathcal{D})\overset{\text{def}}{=} \dfrac{\mathrm{d}\hat{\mathbf{w}}_{\epsilon;z'}}{\mathrm{d}\epsilon}\bigg|_{\epsilon=0} = -H^{-1}_{\hat{\mathbf{w}}} \nabla_{\mathbf{w}}f(z'; \hat{\mathbf{w}}),
\end{equation}
where $H_{\hat{\mathbf{w}}}$ is the Hessian, $f(\cdot; \hat{\mathbf{w}})$ is the loss function, and $\nabla_{\mathbf{w}}f(\cdot; \hat{\mathbf{w}})$ is the gradient of the loss function. 

\subsubsection{Ensemble Learning}
Ensemble learning combines multiple individual models, denoted as weak learners, together to improve prediction and decision-making~\cite {zhou2012ensemble}. By exploiting complementary knowledge and reducing bias and variance, ensemble methods can achieve higher accuracy and robustness compared to single models~\cite{kuncheva2003measures}. The component models' diversity and competence, training data's size and quality, and ensemble techniques are key factors determining effectiveness~\cite{rokach2010ensemble}. These principles of ensemble learning have also been adapted and applied in various designs for exact unlearning.

\subsection{Machine Unlearning}
\subsubsection{Problem Definition}
Machine unlearning refers to the process of removing the influence of specific training data points on an already trained machine learning model ~\cite{chen2021machine}. 
Formally, given a model with parameters $\mathbf{w}^*$ trained on dataset $\mathcal{D}$ using learning algorithm $\mathcal{A}$, and a subset $\mathcal{D}_f \subseteq \mathcal{D}$ to be removed, the machine unlearning algorithm $\mathcal{U}(\mathcal{A}(\mathcal{D}), \mathcal{D}, \mathcal{D}_f)$ aims to obtain new model with parameters $\mathbf{w}^-$ by removing the effects of $\mathcal{D}_f$ while preserving performance on $\mathcal{D} \setminus \mathcal{D}_f$. 

\subsubsection{Application Scenarios of Machine Unlearning}
Machine unlearning enables selectively removing specific data points from trained ML models. This subtractive capability supports key applications:
\begin{itemize}
	\item Privacy Protection: Machine unlearning helps enforce privacy rights and enhances privacy protection~\cite{fu2021knowledge,garg2020formalizing}. It allows removing users' personal data from trained models to comply with regulations such as GDPR and CCPA, which grant users the right to remove their data ~\cite{Bertram2019,garg2020formalizing}. By removing training data points, machine unlearning also reduces the membership inference attacks that aim to determine if certain data was used in training~\cite{chundawat2023zero}.
	\item Improving Security: Machine learning models can be vulnerable to data poisoning attacks, where adversaries inject malicious data into the training set to manipulate the model's behavior. Machine unlearning improves security by removing poisoned data points, making the system robust against such attacks~\cite{cao2018efficient, chundawat2023zero,shan2022poison}.
	\item Enabling Adaptability: Models trained on static datasets may not adapt well as the underlying data distribution changes over time, resulting in outdated or inaccurate models~\cite{gama2014survey,Mirzasoleiman2017}. Machine unlearning facilitates adaptability by removing outdated or unrepresentative data, keeping the model relevant even as the environment evolves~\cite{cao2015towards}.
\end{itemize}
\subsubsection{Challenges in Machine Unlearning}
Machine unlearning faces challenges from inherent properties of ML models as well as practical implementation issues:
\begin{itemize}
	\item Data Dependencies:
	ML models do not simply analyze data points in isolation. Instead, they synergistically extract complex statistical patterns and dependencies between data points~\cite{serra2018overcoming}. Removing an individual point can disrupt the learned patterns and dependencies, potentially leading to a significant decrease in performance~\cite{cao2015towards,chen2021machine, Schelter2021HedgeCut}.
	\item Model Complexity: Large machine learning models such as deep neural networks can have millions of parameters. Their intricate architectures and nonlinear interactions between components make it hard to interpret the model and locate the specific parameters most relevant to a given data point ~\cite{cao2015towards,lin2023erm}. The lack of transparency into how data influences predictions poses challenges for removing dependencies.
	\item Computational Cost: Most machine unlearning techniques require iterative optimization methods such as gradient descent to adjust parameters after removing data. This incurs a significant computational cost, which grows rapidly as the model and dataset size increase~\cite{cao2015towards, bourtoule2021machine}. The computational demands may exceed the available resources when dealing with large-scale datasets and complex models. 
	\item Privacy Leaks: The unlearning process itself can leak information in multiple ways~\cite{carlini2022privacy}. For instance, statistics, such as the time taken to remove a point, can reveal information about it~\cite{chen2021machine,carlini2022privacy}. Additionally, alterations in accuracy and outputs can allow adversaries to infer the characteristics of removed data~\cite{chen2021machine,ye2022learning}. 
	\item Dynamic Environments: Tracing each data point's influence becomes increasingly difficult as the dataset changes dynamically ~\cite{Mirzasoleiman2017,wang2019measure}. Unlearning can also introduce delays that impede prompt model updates needed for low-latency predictions.
\end{itemize}

\subsubsection{Evaluation Metrics for Machine Unlearning Solutions}
\paragraph{Data Erasure Completeness} 
This metric evaluates how thoroughly the unlearning algorithm makes the model remove the target data. It compares the model's predictions or parameters before and after unlearning to quantify the extent of removing. Various distance or divergence measures can be used to quantify the difference between the two models~\cite{jia2021zest}. Representative measures include $L_2$ distance~\cite{wu2020deltagrad} and Kullback-Leibler (KL)  divergence~\cite{golatkar2020eternal}.

\paragraph{Unlearning Time Efficiency} 
This metric can be assessed by comparing the duration required for naive retraining of the model with the time it takes to perform the unlearning process~\cite{mercuri2022introduction}. This efficiency is key for responsive, real-time applications, highlighting the practical advantage of unlearning over retraining methods~\cite{cao2015towards}. 

\paragraph{Resource Consumption}
This metric assesses the memory usage, power consumption, and storage costs incurred during the unlearning process, gauging machine unlearning solutions' practical viability and scalability.   Efficient unlearning algorithms are characterized by their ability to minimize these resource demands while effectively meeting unlearning goals.

\paragraph{Privacy Preservation}
Certified removal~\cite{guo2020certified} is an important privacy metric for approximate unlearning solutions inspired by differential privacy~\cite{dwork2006differential}. It offers a theoretical assurance that a model, after specific data removal, is indistinguishable from a model never trained on that data. This property implies that an adversary cannot extract information about the removed training data from the model, rendering membership inference attacks on the removed data unsuccessful. This property can be represented in two ways: $\epsilon$-certified removal and its more relaxed version, $(\epsilon, \delta)$-certified removal~\cite{guo2020certified}.

\subsubsection{Implementations of Machine Unlearning}
Machine Unlearning has been implemented across computer vision using Convolutional Neural Networks (CNN), natural language processing via Transformer, and graph data processing with Graph Neural Networks (GNNs). The implementation of machine unlearning involves diverse datasets based on the model type, including image data  (e.g., MNIST~\cite{lecun1998mnist}, CIFAR~\cite{krizhevsky2009learning}, SVHN~\cite{netzer2011reading}, and ImageNet~\cite{deng2009imagenet}), text corpora (e.g., IMDB Review~\cite{maas2011learning}, WIKITEXT-103~\cite{merity2016pointer} and OpenWebText Corpus~\cite{Gokaslan2019OpenWeb}), and graph datasets (e.g., Cora~\cite{sen2008collective}, Citeseer~\cite{giles1998citeseer}, and Pubmed~\cite{pubmed}). 
Researchers~\cite{nguyen2022survey} have compiled implementations and studies related to machine unlearning in an open repository \href{https://github.com/tamlhp/awesome-machine-unlearning}{https://github.com/tamlhp/awesome-machine-unlearning}.

\subsection{Comparison with Other Surveys on Machine Unlearning}
This paper distinguishes itself from existing surveys on machine unlearning, including the work of Qu \textit{et al.}~\cite{qu2023learn}, by offering a more comprehensive and extensive examination of the field. Previous surveys have generally categorized unlearning methodologies based on data and model operations, such as data reorganization and model manipulation, as outlined in the surveys~\cite{xu2023machine,shaik2023exploring}, or have categorized approaches as model-agnostic, model-intrinsic, and data-driven~\cite{nguyen2022survey}. While these classifications offer clear and structured insights into the unlearning methodologies, they implicitly suggest that data-centric and model-centric approaches are mutually exclusive, potentially overlooking their interaction and overlap. For example, operations on data reorganization (such as data pruning and data replacement) can lead to adjustments in model manipulation (such as model replacement and model pruning). This interaction highlights a complex dynamic where changes in one domain directly influence the other.

In contrast, our paper categorizes unlearning methods based on their theoretical foundations, such as influence functions, re-optimization, or gradient updates. This approach acknowledges the interdependencies between data and model adjustments and aims to deepen understanding by examining the principles underlying unlearning strategies.

\subsection{Naive Retraining} \label{secnaive}
Naive retraining, also known as fully retraining and retraining from scratch, is to remove the data point from the training dataset and retrain the model again. It is often used as a baseline to evaluate unlearning techniques.

\begin{definition}[Naive Retraining]
	Given a machine learning algorithm $\mathcal{A}(\cdot)$, training dataset $\mathcal{D}$, and a training data point $z'=(x', y')$ to be removed, naive retraining involves retraining on the modified dataset $\mathcal{D} \setminus z'$. Mathematically, it can be represented as:
	\begin{equation}
		\mathcal{A}(\mathcal{D} \setminus z').
	\end{equation}
\end{definition} 

\begin{figure}
	\centering
	\includegraphics[width=0.9\linewidth]{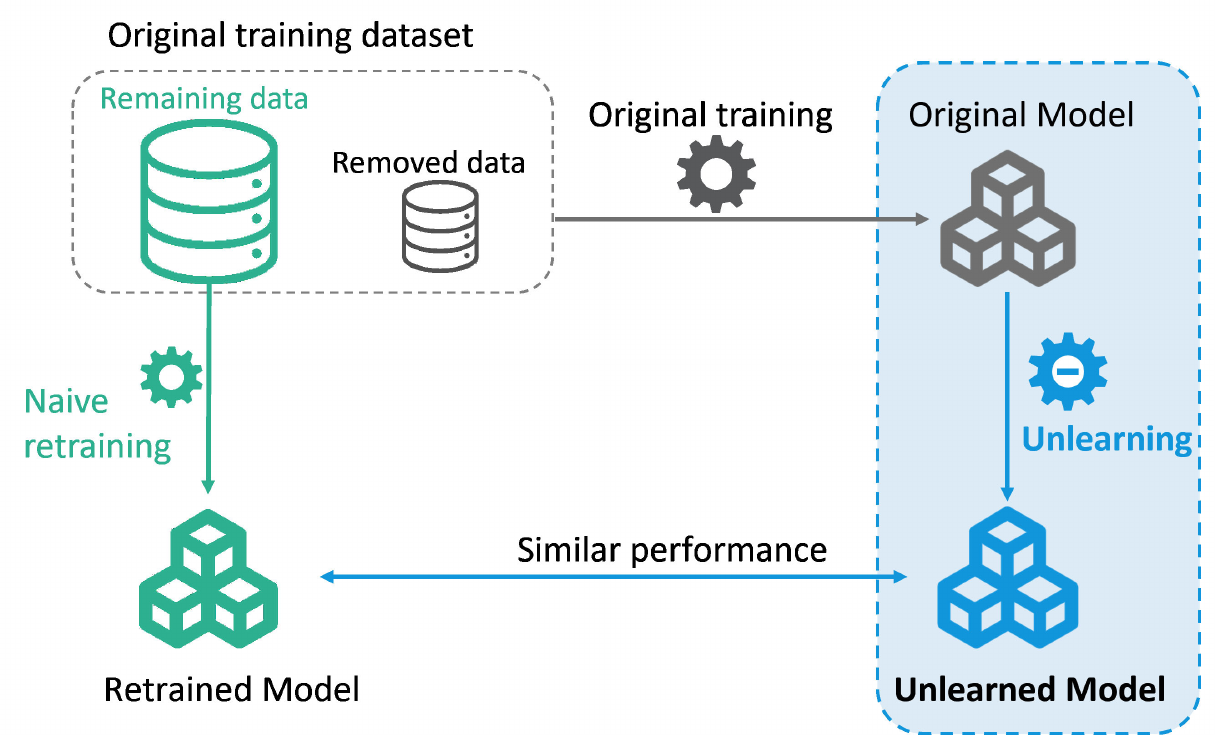}
	\caption{Illustration of Naive Retraining and Machine Unlearning}
	\label{fig:unlearn}
\end{figure}

As shown in Figure \ref{fig:unlearn}, naive retraining first discards the original model and removes the target data from the training dataset. Then, a new model is trained from scratch with the remaining training data. By excluding the data to be removed, the retraining process removes information embedded in the model.

Naive retraining has several drawbacks. It is computationally intensive due to complete parameter re-optimization, time-consuming for complex models and large datasets, and relies on the original training data, which limits its feasibility when access is restricted.

\section{Exact Unlearning} \label{secexact}
Exact unlearning takes a more efficient strategy than naive retraining. This section presents an overview of exact unlearning through the SISA framework, followed by methods based on the SISA framework and other variations of exact unlearning.

\subsection{Overview of Exact Unlearning}
The Sharding, Isolation, Slicing, and Aggregation (SISA)~\cite{bourtoule2021machine} framework is a general approach for exact unlearning. By sharding, isolating, slicing, and aggregating training data, SISA enables targeted data removal without full retraining.  

The key idea of SISA is to divide the training data into multiple disjoint shards, with each shard for training an independent sub-model. The influence of each data point is isolated within the sub-model trained on its shard. When removing a point, only affected sub-models need to be retrained.

\begin{figure}
	\centering
	\includegraphics[width=0.95\linewidth]{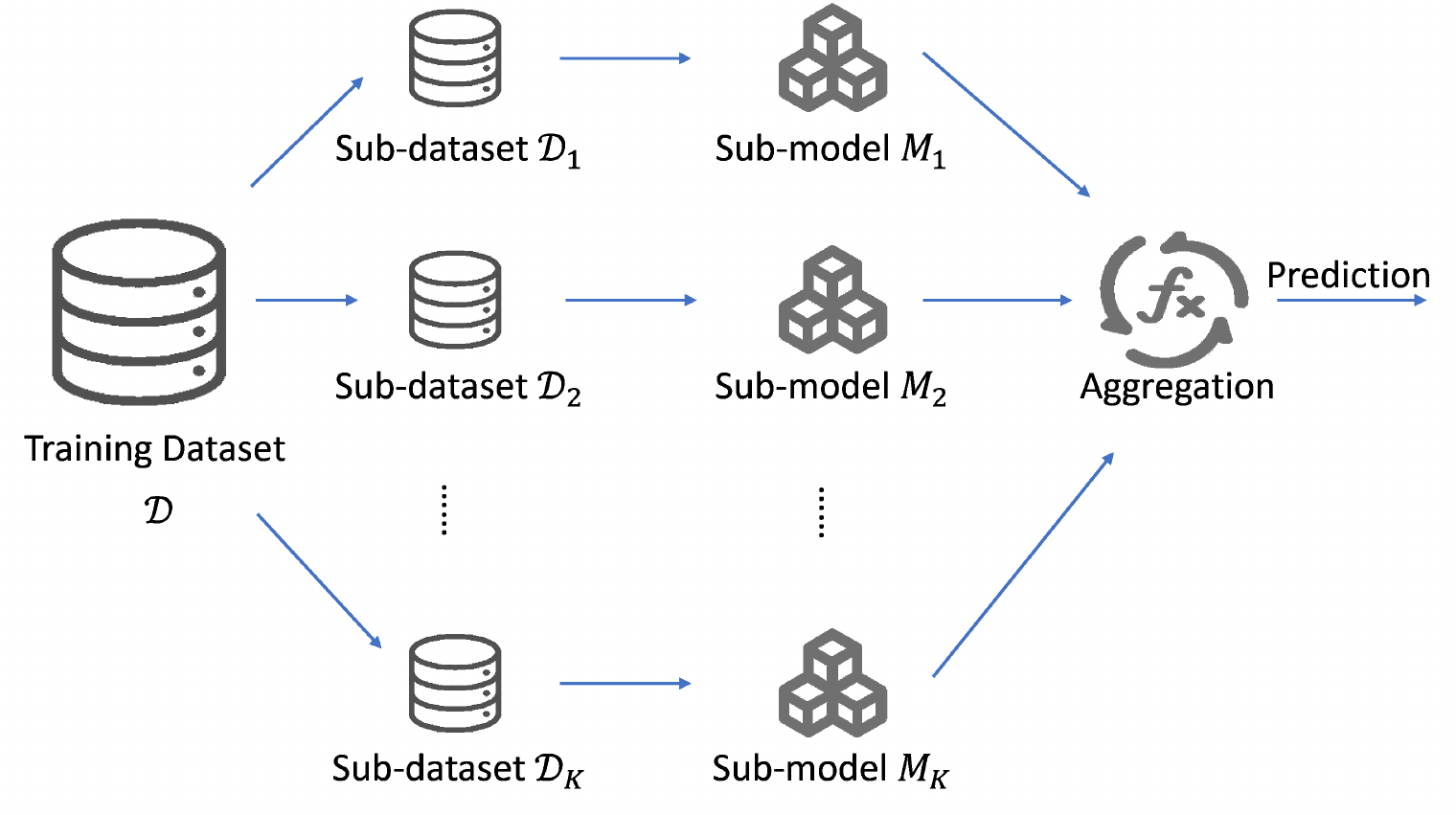}
	\caption{SISA Framework}
	\label{fig:sisa}
\end{figure}

As shown in Figure \ref{fig:sisa}, the implementation of SISA includes four key steps.
\begin{enumerate} [label=(\arabic*)] 
	\item Sharding: The training dataset is divided into multiple disjoint subsets called `shards.' Each shard is used to train a separate sub-model.
	\item Isolation: The sub-models are trained independently of each other, ensuring that the influence of a data point is isolated to the model trained on the shard containing that data point.
	\item Slicing: Within each shard, the data is further divided into `slices'. Models are incrementally trained on these slices. The parameters are stored before including each new slice to track the influence of unlearned data points at a more fine-grained level.
	\item Aggregation: The sub-models trained on each shard are aggregated to form the final model. Aggregation strategies, such as majority voting, allow SISA to maintain good performance.
\end{enumerate}

When unlearning a specific data point, only the sub-models associated with shards containing that data need retraining. The retraining can start from the last parameter saved that does not include the data point to be unlearned. 

SISA offers several advantages over naive retraining. First, it reduces the computational cost and time required for unlearning by training models on smaller shards, retraining only the affected models, and incrementally updating models using slices. Second, it maintains prediction accuracy by aggregating the knowledge of the sub-models. Third, SISA provides a flexible and scalable solution, allowing the system to handle evolving unlearning requests without compromising the model's overall performance.

However, SISA does have limitations. First, the effectiveness of SISA depends on the specific characteristics of the learning algorithm of sub-models and the data. For example, it may not work well for models that learn complex interactions between data points or for data that is not easily divisible into independent shards. Second, SISA requires additional storage resources for keeping separate sub-models and tracking each data point's influence within each slice. Third, the model's generalization ability could be degraded due to isolated training and the tradeoffs involved in aggregation strategies. 

In the next subsection, we discuss some notable improvements and adaptations of SISA for different models, including random forest, graph-based models, and $k$-means. 

\subsection{ Exact Unlearning based on SISA Structure} 
\subsubsection{Exact Unlearning for Random Forest}
Exact unlearning for random forest can be seen as a specific application of the SISA framework. Each tree in the forest is trained on a different subset of data, acting as a shard in the SISA framework. The predictions of the individual trees are aggregated to obtain the final prediction of the random forest. The influence of a data point is isolated within the trees trained on the subset containing that data point. When unlearning a data point, only the trees trained on the relevant subset require retraining.

DaRE~\cite{Brophy2020DaRE} proposes a variant of random forest called Data Removal-Enabled (DaRE) forest. DaRE forest uses a two-level approach with random and greedy nodes in the tree structure. Random nodes, located in upper levels, choose split attributes and thresholds uniformly at random, requiring minimal updates as they are less dependent on data. Greedy nodes in lower levels optimize splits based on criteria such as the Gini index or mutual information. DaRE trees cache statistics at each node and training data at each leaf, allowing for efficient updates of only necessary subtrees when data removal requests are received. This caching and use of randomness improve the efficiency of unlearning.

HedgeCut~\cite{Schelter2021HedgeCut} focuses on unlearning requests with low latency in extremely randomized trees. It introduces the concept of \textit{split robustness} to identify split decisions that may change with removed data. HedgeCut maintains \textit{subtree variants} for such cases, and when unlearning a data point, it replaces the corresponding split with the prepared subtree variants. This operation is quick and straightforward, ensuring a short delay in the unlearning process.

\subsubsection{ Exact Unlearning for Graph-based Model}
The interconnected structure of graph data makes it challenging for graph-based model unlearning, as influence from any single data point spreads across the entire graph. This necessitates the development of specialized graph-based unlearning methods. Exact unlearning for graph-based models aims to efficiently and accurately remove the influence of individual data points on model predictions while accounting for the unique characteristics of graph-structured data.

GraphEraser~\cite{Chen2022GraphEraser} and RecEraser~\cite{Chen2022RecEraser} extend the SISA framework to graph data but use different partitioning and aggregation strategies. GraphEraser is designed for GNNs unlearning. It consists of three phases: balanced graph partition, shard model training, and shard model aggregation. The graph partition algorithms focus on preserving the graph structural information and balancing the shards resulting from the graph partition. The learning-based aggregation method optimizes the importance score of the shard models to improve the global model utility. When a node needs to be unlearned, GraphEraser removes the node from the corresponding shard and retrains the shard model.

While GraphEraser handles general graph data, it may be less optimal for recommendation systems, where collaborative information across users and items is crucial. RecEraser~\cite{Chen2022RecEraser} is specialized for recommendation tasks, where user-item interactions are represented in graphs. It extends the SISA framework by proposing three data partition methods based on users, items, and interactions to divide training data into balanced shards. RecEraser uses an adaptive aggregation method to combine the predictions of the sub-models. This considers both the local collaborative information captured by each sub-model and the global collaborative information captured by all sub-models. Upon receiving a data unlearning request, the affected sub-model and aggregation need retraining in RecEraser. Consequently, RecEraser can make accurate recommendations after unlearning user-item interactions compared to the static weight of sub-models in GraphEraser. 

\subsubsection{Exact Unlearning for $k$-means}
DC-$k$-means~\cite{Ginart2019kmeans} adopts the SISA framework but uses a tree-like hierarchical aggregation method. The training data is randomly divided into multiple subsets, each represented by a leaf node in a perfect $w$-ary tree of height $h$. A $k$-means model is trained on each subset, with each leaf node corresponding to a $k$-means model. The final model is an aggregation of all the $k$-means models at the leaf nodes of the tree, achieved by recursively merging the results from the leaf nodes to the root. To unlearn a data point, the relevant leaf node is located, and the corresponding $k$-means model is updated to exclude that data point. The updated model then replaces the old model at the leaf node, and the changes propagate up the tree to update the final aggregated model. 
\subsubsection{Exact Unlearning for Federated Learning (FL)}
KNOT~\cite{su2023asynchronous} adopts the SISA framework for client-level asynchronous federated unlearning during training. A clustered aggregation mechanism divides clients into multiple clusters. The server only performs model aggregation within each cluster, while different clusters train asynchronously. When a client requests to remove its data, only clients within the same cluster need to be retrained, while other clusters are unaffected and can continue training normally. To obtain an optimal client-cluster assignment, KNOT formulates it as a lexicographic minimization problem. The goal is to minimize the match rating between each client and assigned cluster, considering both training speed and model similarity. This integer optimization problem can be efficiently solved as a Linear Program (LP) using an off-the-shelf LP solver.
\subsubsection{Improvements of SISA}
ARCANE~\cite{Yan2022ARCANE} is designed to overcome the limitations of SISA, aiming to accelerate the exact unlearning process and ensure retrained model accuracy. Unlike SISA's random and balanced data partition, ARCANE divides the dataset into class-based subsets, training sub-models independently using a one-class classifier. This approach reduces accuracy loss by confining unlearning influence to a single class. Besides, ARCANE  introduces data preprocessing methods to reduce retraining costs, including representative data selection, model training state saving, and data sorting by erasure probability. Representative data selection removes redundancies and focuses on selecting the most informative subset of the training set. Training state saving allows for the reuse of previous calculation results, further improving efficiency. Sorting the data by erasure probability enhances the speed of handling unlearning requests.

\begin{table*}[htbp]
	\caption{Exact Unlearning Methods based on SISA}
	\resizebox{\textwidth}{!}{
		\begin{tabular}{m{1.2cm}m{2cm}m{4cm}m{4cm}m{4cm} }
			\hline
			\textbf{Paper} & \textbf{Goal} & \textbf{Design Ideas} & \textbf{Strengths} & \textbf{Weaknesses} \\
			\hline
			\begin{tabular}{@{}c@{}}SISA\\~\cite{bourtoule2021machine} \end{tabular}& Efficient unlearning for general models & 
			\begin{tabitemize}[leftmargin=*, noitemsep, topsep=0pt, partopsep=0pt]
				\item Sharded, isolated, sliced, aggregated training
			\end{tabitemize} & 
			\begin{tabitemize}[leftmargin=*, noitemsep, topsep=0pt, partopsep=0pt]
				\item Applicable to any model
				\item Improves unlearning efficiency
				\item Scalable 
			\end{tabitemize} & 
			\begin{tabitemize}[leftmargin=*, noitemsep, topsep=0pt, partopsep=0pt]
				\item Breaking data dependencies
				\item Additional storage cost for model parameters
				\item Accuracy degradation
			\end{tabitemize} \\
			\hline
			\begin{tabular}{@{}c@{}}DaRE\\~\cite{Brophy2020DaRE} \end{tabular}& Efficient unlearning for random forest & 
			\begin{tabitemize}[leftmargin=*, noitemsep, topsep=0pt, partopsep=0pt]
				\item Use cached statistics to only retrain affected subtrees
				\item Consider subsets of thresholds per attribute
			\end{tabitemize} & 
			\begin{tabitemize}[leftmargin=*, noitemsep, topsep=0pt, partopsep=0pt]
				\item Achieves efficiency by retraining affected subtrees
				\item Analyzes the effect of layers and thresholds 
			\end{tabitemize} & 
			\begin{tabitemize}[leftmargin=*, noitemsep, topsep=0pt, partopsep=0pt]
				\item Additional storage cost for caching statistics
				\item Worst case performance no better than retraining
			\end{tabitemize} \\
			\hline
			\begin{tabular}{@{}c@{}}HedgeCut\\~\cite{Schelter2021HedgeCut} \end{tabular}& Low-latency unlearning for ensemble of random forest& 
			\begin{tabitemize}[leftmargin=*, noitemsep, topsep=0pt, partopsep=0pt]
				\item Use split robustness to identify splitting decisions
				\item Prepare subtree variants for non-robust splits
			\end{tabitemize} & 
			\begin{tabitemize}[leftmargin=*, noitemsep, topsep=0pt, partopsep=0pt]
				\item Low-latency unlearning 
				\item Predictive accuracy similar to random forest
			\end{tabitemize} & 
			\begin{tabitemize}[leftmargin=*, noitemsep, topsep=0pt, partopsep=0pt]
				\item Rely on the accuracy of predicted unlearning requests
				\item Storage cost for subtree variants
			\end{tabitemize} \\
			\hline
			\begin{tabular}{@{}c@{}}DC-$k$-means\\~\cite{Ginart2019kmeans} \end{tabular} & Efficient unlearning for $k$-means& 
			\begin{tabitemize}[leftmargin=*, noitemsep, topsep=0pt, partopsep=0pt]
				\item Divide-and-Conquer approach
				\item Hierarchical aggregation
			\end{tabitemize} & 
			\begin{tabitemize}[leftmargin=*, noitemsep, topsep=0pt, partopsep=0pt]
				\item Theoretical guarantees on removal efficiency
				\item Negligible quality loss
			\end{tabitemize} & 
			\begin{tabitemize}[leftmargin=*, noitemsep, topsep=0pt, partopsep=0pt]
				\item State storage cost 
				\item High removing time complexity in high dimensions
			\end{tabitemize} \\
			\hline
			\begin{tabular}{@{}c@{}}GraphEraser\\~\cite{Chen2022GraphEraser} \end{tabular}& Efficient unlearning for for GNNs& 
			\begin{tabitemize}[leftmargin=*, noitemsep, topsep=0pt, partopsep=0pt]
				\item Community detection for balanced sharding 
				\item Shard model importance scores
			\end{tabitemize} & 
			\begin{tabitemize}[leftmargin=*, noitemsep, topsep=0pt, partopsep=0pt]
				\item Maintains graph structure
				\item Balanced partitioning ensures unlearning efficiency
			\end{tabitemize} & 
			\begin{tabitemize}[leftmargin=*, noitemsep, topsep=0pt, partopsep=0pt]
				\item Additional cost of maintaining massive shards.
				\item Not satisfy the adaptive setting
			\end{tabitemize} \\
			\hline
			\begin{tabular}{@{}c@{}}RecEraser\\~\cite{Chen2022RecEraser} \end{tabular}& Efficient unlearning for recommender systems& 
			\begin{tabitemize}[leftmargin=*, noitemsep, topsep=0pt, partopsep=0pt]
				\item Different data partition strategies
				\item Attention-based adaptive aggregation method
			\end{tabitemize} & 
			\begin{tabitemize}[leftmargin=*, noitemsep, topsep=0pt, partopsep=0pt]
				\item Efficient unlearning with balanced data partition
				\item Good global model utility with adaptive aggregation
			\end{tabitemize} & 
			\begin{tabitemize}[leftmargin=*, noitemsep, topsep=0pt, partopsep=0pt]
				\item Specific to recommendation task
				\item Open problem in the batch setting
			\end{tabitemize} \\
			\hline
			\begin{tabular}{@{}c@{}}KNOT~\cite{su2023asynchronous} \end{tabular}& Federated unlearning during training& 
			\begin{tabitemize}[leftmargin=*, noitemsep, topsep=0pt, partopsep=0pt]
				\item Cluster clients and perform aggregation within clusters
				\item Clusters train asynchronously
			\end{tabitemize} & 
			\begin{tabitemize}[leftmargin=*, noitemsep, topsep=0pt, partopsep=0pt]
				\item Limits retraining to a cluster
				\item Asynchrony allows unaffected clusters to continue training
			\end{tabitemize} & 
			\begin{tabitemize}[leftmargin=*, noitemsep, topsep=0pt, partopsep=0pt]
				\item Performance relies on cluster assignments
				\item Asynchrony can negatively impact model accuracy
			\end{tabitemize} \\\hline
			\begin{tabular}{@{}c@{}}ARCANE\\~\cite{Yan2022ARCANE} \end{tabular}& Efficiently and accurately unlearning & 
			\begin{tabitemize}[leftmargin=*, noitemsep, topsep=0pt, partopsep=0pt]
				\item Multiple one-class classification tasks
				\item Data preprocessing
			\end{tabitemize} & 
			\begin{tabitemize}[leftmargin=*, noitemsep, topsep=0pt, partopsep=0pt]
				\item Maintains accuracy even with large unlearning requests 
				\item Data preprocessing accelerates unlearning
			\end{tabitemize} & 
			\begin{tabitemize}[leftmargin=*, noitemsep, topsep=0pt, partopsep=0pt]
				\item Only applies to supervised models
				\item Rely on preprocessing for efficiency
			\end{tabitemize} \\\hline
		\end{tabular}
	}
	\label{table:exctcomparison}
\end{table*}

\subsection{Non-SISA Exact Unlearning}
Cao \textit{et al.}~\cite{cao2015towards} were inspired by statistical query learning and proposed an intermediary layer called `summations' to decouple machine learning algorithms from the training data. Instead of directly querying the data, learning algorithms rely on these summations. This allows the removal of specific data points by updating the summations and computing the updated model. The unlearning process involves two steps. First, the feature set is updated by excluding the removed data point and re-scoring the features. The updated feature set is generated by selecting the top-scoring features. This process is more efficient than retraining as it does not require examining each data point for each feature. Second, the model is updated by removing the corresponding data if a feature is removed from the feature set or computing the data if a feature is added. Simultaneously, summations dependent on the removed data point are updated, and the model is adjusted accordingly. 

Liu \textit{et al.} ~\cite{liu2022right} propose a rapid retraining approach for FL. When a client requests data removal, all clients perform local data removal, followed by a retraining process on the remaining dataset. This process utilizes a first-order Taylor approximation technique based on the Quasi-Newton method and a low-cost Hessian matrix approximation method, effectively reducing computational and communication costs while maintaining model performance.

\subsection{Comparisons and Discussions}
The comparison of different exact unlearning methods is shown in Table~\ref{table:exctcomparison}. We evaluate their strengths and weaknesses in terms of storage cost, assumptions, model utility, computational cost, scalability, and practicality. Our goal is to provide a clear assessment to help guide future research and select suitable unlearning methods for different applications. 
\begin{enumerate} [label=(\arabic*)] 
	\item Additional Storage Cost: Exact unlearning methods rely on substantial additional storage required for caching model parameters, statistics, or intermediate results. For instance, SISA requires the storage of model parameters for each shard and slice, while HedgeCut requires the storage of subtree variants. This hinders scalability to large models or frequent unlearning requests. Developing low-storage approaches would be beneficial.
	\item Strong Assumptions: Some methods make strong assumptions about the learning algorithm or data characteristics. SISA may struggle with highly dependent data, while statistical query learning requires algorithms to be expressed in a summation form. Methods tailored to specific models, such as DaRE and HedgeCut, GraphEraser, and RecEraser, have limited applicability. More flexible and generalizable techniques are needed. 
	\item Model Utility: Most methods claim they maintain accuracy after unlearning but lack thorough analysis across diverse settings. Rigorous evaluations of different models, datasets, and removal volumes are imperative to provide concrete utility guarantees.
	\item Computational Cost: Exact unlearning methods add computational costs during initial training because of multiple sub-models training and aggregation. It may not be feasible when computational resources are limited. 
	\item Managing Evolving Data: Existing methods focus on removing data in fixed training sets. Handling dynamically changing data with continuous insertion and removal requests remains an open problem.
\end{enumerate}

In summary, while existing exact unlearning methods enable efficient and accurate data removal, they have limitations related to storage, assumptions, utility maintenance, and scalability. The most suitable method depends on the application's specific requirements, including the data type, model type, available resources, and the desired balance between efficiency and accuracy. Further research should aim to develop generally applicable techniques with low cost that provably sustain model accuracy even after repeated unlearning.  

\section{Approximate Unlearning}\label{secapp}
Approximate unlearning aims to minimize the influence of unlearned data to an acceptable level while achieving an efficient unlearning process. It has several advantages over exact unlearning techniques, including better computational efficiency, less storage cost, and higher flexibility.
\begin{itemize}
	\item Computational Efficiency: Exact unlearning methods require retraining at the algorithm level using remaining data, which can be computationally expensive, particularly for large datasets. In contrast, approximate unlearning focuses on minimizing the data's influence to be removed rather than completely removing it. For example, Guo \textit{et al.}~\cite{guo2020certified} proposed a method that adjusts the model parameters based on the calculated influence of the removed data. This approach is less computationally intensive than the full re-computation required in exact unlearning.
	\item Storage Overhead: Exact unlearning typically requires storing the training dataset and multiple sub-models, leading to high storage costs. Conversely, approximate unlearning, such as the one proposed by Sekhari \textit{et al.}~\cite{sekhari2021remember}, only requires storing cheap-to-compute data statistics. As a result, the storage burden associated with exact unlearning is significantly alleviated.
	\item Flexibility: Exact unlearning methods are often tailored to specific learning models or data structures, limiting their applicability. On the other hand, many approximate unlearning are more model-agnostic. They can be applied to diverse learning algorithms without requiring specific model or data structure modifications. This enhanced flexibility allows approximate unlearning to be more widely applicable compared to exact unlearning.
\end{itemize}

It is important to note that approximate unlearning represents a tradeoff between unlearning completeness and unlearning efficiency~\cite{cao2015towards}. In some cases, a slight decrease in completeness, but a significant speed-up of the unlearning process and savings in computation and storage costs, is an acceptable tradeoff. By carefully considering these tradeoffs, approximate unlearning provides an effective and practical approach for adapting models to new data and tasks.

\subsection{Overview of Approximate Unlearning}
Approximate unlearning is a process that aims to minimize the influence of unlearned data to an acceptable level rather than completely removing it. It involves the following key steps.
\begin{enumerate} [label=(\arabic*)] 
	\item Computation of Influence: Calculate the influence of the data points that need to be unlearned on the original model. This involves determining how these data points affect the model.
	\item Adjustment of Model Parameters: Modify the model parameters to reverse the influence of the data being removed. This adjustment typically involves methods such as reweighting or recalculating optimal parameters and modifying the model so that it behaves as if it was trained on the dataset without the unlearned data points.
	\item Addition of Noise (Optional): Carefully calibrated noise is added to prevent the removed data from being inferred from the updated model. This step ensures the confidentiality of the training dataset.
	\item Validation of Updated Model: Evaluate the performance of the updated model to ensure its effectiveness. This validation step may involve cross-validation or testing on a hold-out set to assess the model's accuracy and generalization.
\end{enumerate}

By following these steps, approximate unlearning efficiently reduces the influence of specific data points in a trained model. This approach provides a practical alternative to exact unlearning, particularly in scenarios where computational cost, storage cost, and flexibility are crucial factors.

The subsequent subsections provide more details on specific approximate unlearning. Section \ref{secif} discusses methods based on the influence function, while Sections \ref{secreopt} and \ref{secgd} explore methods based on re-optimization and gradient update, respectively. Section \ref{secgraph} introduces methods specifically designed for graph data, and Section \ref{secother} covers other notable methods in the field. Each section provides a detailed summary of the respective methods, discussing their applications, advantages, and limitations.

\subsection{Approximate Unlearning based on Influence Function of the Removed Data} \label{secif}
This section explores representative works leveraging influence functions for approximate unlearning and discusses their applications and limitations. The core idea shared by these representative works is to use the influence function to estimate data points' influence for removal. Some works ~\cite{guo2020certified,sekhari2021remember, suriyakumar2022algorithms, mehta2022deep,wu2022puma,tanno2022repairing} develop unlearning algorithms that approximately compute the influence function of data points on the model parameters, and then update the parameters to remove the influence of unlearned data points. Some ~\cite{sekhari2021remember, suriyakumar2022algorithms, mehta2022deep} also focus on efficiently approximating the influence functions by considering only relevant parameters or using sampling because the direct computation of influence functions can be computationally expensive.

We start by discussing the pioneering work of Guo \textit{et al.}~\cite{guo2020certified}. Then, we explore how subsequent research has built upon Guo's work to address its limitations, particularly by reducing the computational cost associated with inverting the Hessian matrix. Finally, we discuss the future directions of this research field.

Guo \textit{et al.}\cite{guo2020certified} introduced influence functions for data removal and achieved certified removal of $L_2$-regularized linear models. Specifically, linear models are usually trained using a differentiable convex loss function as shown in Eq.(\ref{certifloss}). 
\begin{equation} \label{certifloss}
	F(\mathcal{D};\mathbf{w})=	\sum_{z\in\mathcal{D}} f(z; \mathbf{w})+\frac{\lambda n}{2}	\left \| \mathbf{w}\right \|^2_2,
\end{equation}
where $f(z; \mathbf{w})$ is a convex loss function. To protect the information of the removed data points, Guo \textit{et al.} propose to add random perturbation~\cite{chaudhuri2011differentially} during the training process to protect the gradient information. Thus, the loss function used for training is as shown in Eq. (\ref{noise}):
\begin{equation}\label{noise}
	F_{\mathbf{b}}(\mathcal{D};\mathbf{w})=	\sum_{z\in\mathcal{D}} f(z; \mathbf{w})+\frac{\lambda n}{2}	\left \| \mathbf{w}\right \|^2_2+\mathbf{b}^{\top}\mathbf{w},
\end{equation} 
where $\mathbf{b}$ is a random vector.
The parameters of the model is $\mathbf{w}^*$, where
\begin{equation}
	\mathbf{w}^*=A(\mathcal{D})=\operatorname{argmin}_{\mathbf{w}} F_{\mathbf{b}}(\mathcal{D}; \mathbf{w} ).
\end{equation}
Suppose the data point $z'=( x', y')$ is to be removed from the training set. The process of Newton update removal mechanism to remove $z'$ is as follows:
\begin{enumerate} [label=(\arabic*)] 
	\item Calculate the influence of the removed data point on the model parameters. The loss gradient at $z'$ is 
	\begin{equation} \label{guograd}
		\Delta = \lambda \mathbf{w}^* + \nabla f(z'; \mathbf{w}).
	\end{equation} 
	According to the influence function~\cite{Koh2017Understanding}, the influence of $z'$ on the original model is $-H_{\mathbf{w}^*}^{-1} \Delta$~\cite{Koh2017Understanding}, where $H_{\mathbf{w}^*}$ is the Hessian of the loss function 
	\begin{equation} \label{guoh}
		H_{\mathbf{w}^*}=\nabla^2 F\left(\mathcal{D}_r;\mathbf{w}^* \right), \mathcal{D}_r=\mathcal{D}\setminus z'.
	\end{equation}
	This one-step Newton update is applied to the gradient influence of the removed point $z'$.
	\item Adjust the model parameters $\mathbf{w}^*$ to removes the influence of the $z'$ from the model. The new model parameters $\mathbf{w}^-$ are given by 
	\begin{equation} \label{guow-}
		\mathbf{w}^- = \mathbf{w}^* + H_{\mathbf{w}^*}^{-1} \Delta.
	\end{equation}
\end{enumerate}

Steps (1) and (2) make the new model parameters $\mathbf{w}^{-}$ approximate the original model parameters $\mathbf{w}^{*}$ as closely as possible while ensuring the removal of specific data points from the original dataset. The process also ensures the performance of the modified model and minimizes the leakage of information about the removed data points by adding random perturbations to the model parameters during training and achieves a certified-removal mechanism.

While Guo \textit{et al.}~\cite{guo2020certified} have provided valuable insights into approximate unlearning, their proposed removal mechanism has limitations. 1) It requires inverting the Hessian matrix, which is computationally expensive. Moreover, inverting the Hessian matrix can be numerically unstable for large models because the condition number of the Hessian matrix grows linearly with the number of model parameters. The large condition number leads to error amplification during matrix inversion, resulting in an inaccurate influence approximation.
2) It fails to work for models with non-convex losses because the influence function relies on the convexity assumption to ensure the estimated influence accurately reflects the true influence of removed points. For non-convex models, the loss surface is much more complex, so the estimated influence may not correlate well with the actual impact.
3) There remains a large gap between the data-dependent bound and the true gradient residual norm after unlearning. This gap indicates looseness in the analysis, which needs tightening to provide a practically useful certified bound on the degree of removal.

Building upon Guo's work, Sekhari \textit{et al.}~\cite{sekhari2021remember} does not require full access to the training dataset during the unlearning process. By using cheap-to-store data statistics $\nabla^2 \widehat{F}(\mathcal{D}; \mathbf{w}^*)$ as shown in Eq.(\ref{remh}), they enable efficient unlearning without the need for the entire training data reducing computational and storage requirements, in contrast to Eq.(\ref{guoh}) and Eq.(\ref{guow-}).
\begin{equation} \label{remh}
	\begin{split}
		&\widehat{H}=\frac{1}{n-m}\left(n \nabla^2 \widehat{F}(\mathcal{D}; \mathbf{w}^*)-\sum_{z' \in \mathcal{D}_f} \nabla^2 f(z';\mathbf{w}^*)\right),\\ 
		&\mathbf{w}^-=\mathbf{w}^*+\frac{1}{n-m}(\widehat{H})^{-1} \sum_{z' \in \mathcal{D}_f} \nabla f(z';\mathbf{w}^*) .
	\end{split}
\end{equation}
They also emphasize the importance of test loss and add noise after adjusting model parameters to ensure model performance. This ensures privacy protection without compromising the accuracy and performance of the model. 

Suriyakumar \textit{et al.}~\cite{suriyakumar2022algorithms} propose a more computationally efficient algorithm for online data removal from models trained with Empirical Risk Minimization (ERM). This improvement is achieved by using the \textit{infinitesimal jackknife}, a technique that approximates the influence of excluding a data point from the training dataset on the model parameters. This avoids the need to compute and invert a different Hessian matrix for each removal request, which was required by prior methods~\cite{guo2020certified, sekhari2021remember}. Their approach enables efficient processing of online removal requests while maintaining similar theoretical guarantees on model accuracy and privacy. Moreover, by integrating the infinitesimal jackknife with Newton methods, their algorithm can accommodate ERM-trained models with non-smooth regularizers, broadening applicability.

Mehta \textit{et al.} ~\cite{mehta2022deep} improve the efficiency of Hessian matrix inversion in deep learning models. They introduce a selection scheme, L-CODEC, which identifies a subset of parameters to update, removing the need to invert a large matrix. This avoids updating all parameters, focusing only on influential ones. Building on this, they propose L-FOCI to construct a minimal set of influential parameters using L-CODEC incrementally. Once the subset of parameters to update is identified, they apply a blockwise Newton update to the subset. By focusing computations only on influential parameters, their approach makes approximate unlearning feasible for previously infeasible large deep neural networks.

Unlike the approach by Guo \textit{et al.}~\cite{guo2020certified}, which only adjusts the linear decision-making layer of a model, PUMA~\cite{wu2022puma} modifies all trainable parameters, offering a more thorough solution to data removal. The main purpose of PUMA is to maintain the model's performance after data removal, rather than just monitoring whether the modified model can produce similar predictions to a model trained on the remaining data, as Guo \textit{ et al.}'s method does. To achieve this, PUMA uses the influence function to measure the influence of each data point on the model's performance and then adjusts the weight of the remaining data to compensate for the removal of specific data points. 

Tanno \textit{et al.}~\cite{tanno2022repairing} propose a Bayesian continual learning approach to identify and erase detrimental data points in the training dataset. They use influence functions to measure the influence of each data point on the model's performance, allowing them to identify the most detrimental training examples that have caused observed failure cases. The model is updated to erase the influence of these points by approximating a `counterfactual' posterior distribution, where the harmful data points are assumed to be absent. The authors propose three methods for updating the model weights, one of which is a variant of the Newton update proposed by Guo \textit{et al.}~\cite{guo2020certified}.  

Warnecke \textit{et al.} ~\cite{warnecke2023machine} point out that unlearning should not be limited to removing entire data points. Instead, it should enable corrections at various granularities within the training data. They propose a method that uses influence functions to remove specific features and labels from a trained model. By reformulating influence estimation as a form of removing, the authors derive an approach that maps changes in training data retrospectively to closed-form updates of model parameters. These updates can be computed efficiently even when large portions of the training data are affected, effectively correcting the problematic features and labels within the model.

\begin{table*}[htbp]
	\centering
	\caption{Approximate Unlearning based on Influence Function} \label{tbif}
	\resizebox{\textwidth}{!}{
		\begin{tabular}{m{1.5cm}m{2cm}m{3.5cm}m{3.5cm}m{4cm} }
			\hline
			\textbf{Paper} & \textbf{Goal} & \textbf{Design Ideas} & \textbf{Strengths} & \textbf{Weaknesses} \\ \hline
			\begin{tabular}{@{}l@{}}Guo \textit{et al.} \\ (2020)~\cite{guo2020certified} \end{tabular}
			& Certified removal & 
			\begin{tabitemize}[leftmargin=*,noitemsep,topsep=0pt,partopsep=0pt]
				\item One-step Newton update/influence function
				\item Difference privacy
			\end{tabitemize}
			& 
			\begin{tabitemize}[leftmargin=*,noitemsep,topsep=0pt,partopsep=0pt]
				\item Efficient training data removal
				\item Strong theoretical certified removal guarantee
			\end{tabitemize}
			& 
			\begin{tabitemize}[leftmargin=*,noitemsep,topsep=0pt,partopsep=0pt]
				\item Relies on convexity
				\item High computational cost for inverting the Hessian matrix
			\end{tabitemize}
			\\
			\hline
			\begin{tabular}{@{}l@{}}Sekhari \textit{et al. }\\ (2021)~\cite{sekhari2021remember} \end{tabular} & Efficient unlearning with generalization guarantees & 
			\begin{tabitemize}[leftmargin=*,noitemsep,topsep=0pt,partopsep=0pt]
				\item Use influence functions to identify important data points
				\item Store cheap data statistics
			\end{tabitemize}
			& 
			\begin{tabitemize}[leftmargin=*,noitemsep,topsep=0pt,partopsep=0pt]
				\item Considers test loss instead of just training loss
				\item Reduce computational and storage costs
				\item Give the deletion capacity
			\end{tabitemize}
			& 
			\begin{tabitemize}[leftmargin=*,noitemsep,topsep=0pt,partopsep=0pt]
				\item Relies on convexity
				\item Relies on storage of data statistics
				\item Does not handle finite/discrete hypothesis classes
			\end{tabitemize}
			\\
			\hline
			\begin{tabular}{@{}l@{}}Suriyakumar\\ \textit{et al. } (2022)\\~\cite{suriyakumar2022algorithms} \end{tabular}
			& Efficient unlearning for ERM models & 
			\begin{tabitemize}[leftmargin=*,noitemsep,topsep=0pt,partopsep=0pt]
				\item Infinitesimal jackknife 
				\item Newton update
			\end{tabitemize}& 
			\begin{tabitemize}[leftmargin=*,noitemsep,topsep=0pt,partopsep=0pt]
				\item Computationally efficient online unlearning
				\item Accommodating non-smooth regularizers
			\end{tabitemize}
			& 
			\begin{tabitemize}[leftmargin=*,noitemsep,topsep=0pt,partopsep=0pt]
				\item Specific to ERM models
				\item Inefficient for batch removal
			\end{tabitemize}
			\\
			\hline
			\begin{tabular}{@{}l@{}}Mehta \textit{et al. }\\ (2022)~\cite{mehta2022deep} \end{tabular}
			& Efficient unlearning for DNN & 
			\begin{tabitemize}[leftmargin=*,noitemsep,topsep=0pt,partopsep=0pt]
				\item Conditional independence-based parameter selection
			\end{tabitemize}
			& 
			\begin{tabitemize}[leftmargin=*,noitemsep,topsep=0pt,partopsep=0pt]
				\item Avoids full Hessian inverse
				\item Improves unlearning efficiency in DNN
			\end{tabitemize}
			& 
			\begin{tabitemize}[leftmargin=*,noitemsep,topsep=0pt,partopsep=0pt]
				\item Relies on weighted sampling based on Lipschitz constants of filters/layers
				\item Dependence on newly developed optimization tools
			\end{tabitemize}
			\\
			\hline
			PUMA~\cite{wu2022puma} & Maintain performance during unlearning & 
			\begin{tabitemize}[leftmargin=*,noitemsep,topsep=0pt,partopsep=0pt]
				\item Performance unchanged model augmentation
			\end{tabitemize}
			& 
			\begin{tabitemize}[leftmargin=*,noitemsep,topsep=0pt,partopsep=0pt]
				\item Maintain performance after removal
				\item Computationally efficient
			\end{tabitemize}
			& 
			\begin{tabitemize}[leftmargin=*,noitemsep,topsep=0pt,partopsep=0pt]
				\item Limited evaluation on simple datasets
				\item Sensitive to hyperparameters
			\end{tabitemize}
			\\
			\hline
			\begin{tabular}{@{}l@{}}Tanno \textit{et al. }\\ (2022)~\cite{tanno2022repairing}\end{tabular}
			& Repair model by data removal & 
			\begin{tabitemize}[leftmargin=*,noitemsep,topsep=0pt,partopsep=0pt]
				\item Identify detrimental data via influence function
				\item Remove via posterior approximation
			\end{tabitemize}
			& 
			\begin{tabitemize}[leftmargin=*,noitemsep,topsep=0pt,partopsep=0pt]
				\item Model-agnostic framework 
				\item Identify causes of failures
			\end{tabitemize}
			& 
			\begin{tabitemize}[leftmargin=*,noitemsep,topsep=0pt,partopsep=0pt]
				\item Limited to detrimental data removal 
				\item Cause identification can be computationally intensive
			\end{tabitemize}
			\\
			\hline
			\begin{tabular}{@{}l@{}}Warnecke \textit{et al.}\\ (2023)~\cite{warnecke2023machine} \end{tabular}
			& Unlearn features and labels & 
			\begin{tabitemize}[leftmargin=*,noitemsep,topsep=0pt,partopsep=0pt]
				\item Use influence functions as updates for features or labels
			\end{tabitemize}
			& 
			\begin{tabitemize}[leftmargin=*,noitemsep,topsep=0pt,partopsep=0pt]
				\item Effective unlearning of features/labels
				\item Efficient closed-form updates
			\end{tabitemize}
			& 
			\begin{tabitemize}[leftmargin=*,noitemsep,topsep=0pt,partopsep=0pt]
				\item Efficacy drops as affected features/labels increase
				\item No guarantee for non-convex models
			\end{tabitemize}
			\\
			\hline
		\end{tabular}
	}
\end{table*}

\textbf{\textit{Comparisons and Discussions. } }
The influence function was first introduced for efficient data removal by Guo \textit{et al.} ~\cite{guo2020certified}, providing a one-step Newton update to remove data points based on their influence on model parameters. However, this pioneering work relied on convexity assumptions and suffered from high computational costs due to the need to invert the Hessian matrix. Subsequent research addressed these limitations by developing more efficient approximations of influence functions. The summary and comparison of approximate unlearning based on influence functions are in Table~\ref{tbif}.

A key challenge in this field is the computational cost associated with inverting the Hessian matrix, a step necessary for estimating the influence of data points and updating model parameters. Several strategies have been proposed to address this issue.
Sekhari \textit{et al.}~\cite{sekhari2021remember} reduced storage and computation costs by avoiding the need for the full training data. Suriyakumar \textit{et al.} ~\cite{suriyakumar2022algorithms} proposed using the infinitesimal jackknife technique to efficiently approximate influence in an online setting, also extending unlearning to non-smooth regularizers. Mehta \textit{et al.}~\cite{mehta2022deep} introduced a conditional independence-based selection of influential parameters to avoid inverting large matrices, enabling unlearning in deep neural networks. These techniques demonstrate the potential for efficient approximate unlearning in practical scenarios.

Moreover, considerations such as test loss~\cite{sekhari2021remember}, thoroughness of removal~\cite{wu2022puma}, and finer-grained corrections~\cite{warnecke2023machine} indicate that revised algorithms are becoming increasingly suitable for diverse real-world applications. The extension of these methods to non-convex models~\cite{suriyakumar2022algorithms,warnecke2023machine} and deep neural networks~\cite{mehta2022deep,warnecke2023machine} further highlights their potential.

However, significant challenges remain to be addressed. The accuracy and efficiency of influence estimation and parameter updating require analysis and could potentially be improved. Furthermore, the optimal selection of parameters for updates and the best techniques for influence estimation remain open questions. Connections to differential privacy and information theory may yield valuable insights into inherent limits.

In summary, approximate unlearning based on influence functions shows promise for efficient data removal. This direction enables important progress on algorithmic data removal and its impacts. With continued research, it is expected to play a key role in addressing privacy regulation challenges in the era of big data.

\subsection{Approximate Unlearning based on Re-optimization after Removing the Data}\label{secreopt}
The core idea of approximate unlearning based on re-optimization is to iteratively adjust a model to effectively forget specific data points while maintaining overall performance. The key steps are:
\begin{enumerate} [label=(\arabic*)] 
	\item Train a model $\mathcal{M}(\mathbf{x}; \mathbf{w})$ with parameters $\mathbf{w}$ on the full dataset $\mathcal{D}$. The original loss function is $F$, and the minimum value is obtained at $\mathbf{w}^*$. 
	\item Define a loss function $F(\mathcal{D}_r;\mathbf{w})$ that maintain accuracy on remaining data $\mathcal{D}_r$ while removing information about data to be forgotten $\mathcal{D}_f$.
	\item Re-optimize the model from $\mathbf{w}^*$ by finding updated parameters $\mathbf{w}^-$ that minimize $F(\mathcal{D}_r;\mathbf{w})$. The updated model $\mathcal{M}(\mathbf{x};\mathbf{w}^-)$ retains performance on $\mathcal{D}_r$ while statistically behaving as if trained without $\mathcal{D}_f$.
\end{enumerate}

Research in this area has proposed different techniques to implement the key steps above. They have adopted different techniques for selective removing/forgetting based on application goals.

Golatkar \textit{et al.}~\cite{golatkar2020eternal} propose an optimal quadratic scrubbing algorithm to achieve selective forgetting in deep networks. Selective forgetting is defined as a process that modifies the network weights using a scrubbing function $S(\mathbf{w})$ to make the distribution indistinguishable from weights of a network never trained on the forgotten data. Selective forgetting is measured by the KL divergence. If the KL divergence between the network weight distribution after selective forgetting and the network weight distribution that has never seen the forgotten data is zero, the two distributions are exactly the same, which indicates complete forgetting. 

The authors assume that the quadratic loss function $L(\mathcal{D};\mathbf{w})$ can be decomposed into:
\begin{equation}
	L(\mathcal{D};\mathbf{w}) = L(\mathcal{D}_f;\mathbf{w}) + L(\mathcal{D}_r;\mathbf{w}),
\end{equation} 
where $L(\mathcal{D}_f;\mathbf{w})$ is the loss of the data to be removed and $L(\mathcal{D}_r;\mathbf{w})$ is the loss of remaining data. 

The goal of optimization is to minimize \textit{Forgetting Lagrangian} $\mathcal{L} $:
\begin{equation}
	\mathcal{L} = \mathbb{E}_{S(\mathbf{w})}[L(\mathcal{D}_r;\mathbf{w})] + \lambda \operatorname{KL}[P(S(\mathbf{w})|\mathcal{D}) || P(S_0(\mathbf{w})|\mathcal{D}_r)],
\end{equation}
where $\mathbb{E}_{S(\mathbf{w})}[L(\mathcal{D}_r;\mathbf{w})]$ is the expected loss on the remaining data, $\lambda$ is a hyperparameter that trades off residual information about the data to be removed and accuracy on the remaining data.

The robust scrubbing function $S_t(\mathbf{w}) $ that efficiently removes information about specific training data is:
\begin{equation}\label{eqscrub}
	\begin{aligned}
		S_t(\mathbf{w}) = \mathbf{w} + e^{-Bt}e^{At}d &+ e^{-Bt}(d - d_r) - d_r \\
		&+ (\lambda\sigma_h^2)^{1/4}B^{-1/4}n,
	\end{aligned}
\end{equation}
where $n \sim N(0, I)$, $A=\nabla^2 L(\mathcal{D};\mathbf{w}), B=\nabla^2 L(\mathcal{D}_r;\mathbf{w}), d=A^{-1} \nabla_\mathbf{w} L(\mathcal{D};\mathbf{w})$ and $d_r=B^{-1} \nabla_\mathbf{w} L(\mathcal{D}_r;\mathbf{w})$. $\lambda$ balances information retention and accuracy. The hyperparameter $\sigma_h$ reflects the error in approximating the Stochastic Gradient Descent (SGD).

Using Fisher Information Matrix (FIM)~\cite{martens2020new} to approximate the Hessian matrix, $S_t(\mathbf{w})$ simplifies to:
\begin{equation}
	S(\mathbf{w}) = \mathbf{w} + (\lambda\sigma_h^2)^{1/4}\mathcal{F}^{-1/4},
\end{equation}
where $\mathcal{F}$ is the FIM computed at $\mathbf{w}$ for $\mathcal{D}_r$.

In their follow-up work~\cite{golatkar2020forgetting}, Golatkar \textit{et al.} note that weight changes may not affect deep network outputs due to overparameterization. Consequently, attackers could still extract removed data $\mathcal{D}_f$ from the outputs. To address this issue, they focus on \textit{final activations}. These activations represent the model's response to input data and more directly reflect memory and removing processes. They use a Neural Tangent Kernel (NTK) to correlate weights and activations and introduce an NTK-based scrubbing process to achieve removing by minimizing the difference between the activation of the network on the removing dataset and the target model. 

Later, Golatkar \textit{et al. } ~\cite{golatkar2021mixed} consider mixed-privacy settings where only some user data needs to be removed, while core data are retained. The key insight is to separate the model into two sets of weights: non-linear core weights and linear user weights. Non-linear core weights are trained conventionally on the core data, ensuring they only contain knowledge from the core data that does not need to be removed. Conversely, the linear user weights are obtained by minimizing a quadratic loss on all user data. To remove a subset of user data, the optimal user weight update is directly computed by minimizing the loss on the remaining user data. This aligns with the theoretical optimal update for quadratic loss functions and achieves efficient, accurate removal in mixed-privacy settings without reducing core data accuracy.

Shibata \textit{ et al.} present Learning with Selective Forgetting (LSF)~\cite{Shibata2021} to achieve class-level selective forgetting in lifelong learning. They introduce a loss function $F$ with four components: classification loss $F_C$ to ensure the accuracy of classification, mnemonic loss $F_M$ tying each class to an embedded code, selective forgetting loss $F_{S F}$ to remove classes marked for removal, and regularization loss $F_R$ to prevent catastrophic forgetting. 
\begin{equation}\label{eqslf}
	F=\overbrace{F_C+F_M}^{\text {new task }}+\overbrace{F_{S F}+F_R}^{\text {previous tasks }} .
\end{equation}
The mnemonic codes allow selective forgetting of any class by discarding its code without the original data. The model can selectively forget certain classes by re-optimizing on new loss $F$ as illustrated in Eq.(\ref{eqslf}) without the mnemonic codes of classes to be forgotten. 

\begin{table*}[htbp]
	\caption{Approximate Unlearning based on Re-optimization}
	\label{tbreopt}
	\resizebox{\textwidth}{!}{
		\begin{tabular}{m{1.5cm}m{2.5cm}m{3.5cm}m{3.5cm}m{3.5cm} }
			\hline
			\textbf{Paper} & \textbf{Goal} & \textbf{Design Ideas} & \textbf{Strengths} & \textbf{Weaknesses} \\ \hline
			\raisebox{0.5\height}{\begin{tabular}{@{}l@{}}Golatkar \textit{et al. }\\ (2020a)~\cite{golatkar2020eternal} \end{tabular}}& Selective forgetting in deep networks &
			\begin{tabitemize}[leftmargin=*,noitemsep,topsep=0pt,partopsep=0pt]
				\item Optimal quadratic scrubbing on weights
				\item Add noise tailored to the loss landscape
			\end{tabitemize}
			& 
			\begin{tabitemize}[leftmargin=*,noitemsep,topsep=0pt,partopsep=0pt]
				\item Formal definition of selective forgetting
				\item Provides upper bound on remaining info
			\end{tabitemize}
			& \begin{tabitemize}[leftmargin=*,noitemsep,topsep=0pt,partopsep=0pt]
				\item Rely on the stability of SGD
				\item 	Computationally expensive, limited scalability 
			\end{tabitemize}
			\\ \hline
			\raisebox{0.5\height}{\begin{tabular}{@{}l@{}}Golatkar \textit{et al. }\\ (2020b)~\cite{golatkar2020forgetting} \end{tabular}}
			& Selective forgetting from input-output observations & 
			\begin{tabitemize}[leftmargin=*,noitemsep,topsep=0pt,partopsep=0pt]
				\item Analysis based on final activations
				\item NTK-based scrubbing
			\end{tabitemize}
			&
			\begin{tabitemize}[leftmargin=*,noitemsep,topsep=0pt,partopsep=0pt]
				\item Provides tighter black-box bounds with limited queries
				\item Handles over-parameterized models
			\end{tabitemize}
			&
			\begin{tabitemize}[leftmargin=*,noitemsep,topsep=0pt,partopsep=0pt]
				\item Relies on linearization assumptions
				\item Computationally expensive
			\end{tabitemize}
			\\ \hline
			\raisebox{0.5\height}{\begin{tabular}{@{}l@{}}Golatkar \textit{et al. }\\ (2021)~\cite{golatkar2021mixed} \end{tabular}}
			& Selective forgetting for large deep networks & 
			\begin{tabitemize}[leftmargin=*,noitemsep,topsep=0pt,partopsep=0pt]
				\item Mixed-privacy setting
				\item User weights obtained by minimizing quadratic loss
			\end{tabitemize}
			&
			\begin{tabitemize}[leftmargin=*,noitemsep,topsep=0pt,partopsep=0pt]
				\item Maintaining accuracy on large vision tasks
				\item Provides information bounds
			\end{tabitemize}
			& 
			\begin{tabitemize}[leftmargin=*,noitemsep,topsep=0pt,partopsep=0pt]
				\item Relies on the strong convexity assumptions
				\item Forgetting quality depends on data subsets
			\end{tabitemize}
			\\ \hline
			\raisebox{0.5\height}{\begin{tabular}{@{}l@{}}Shibata \textit{et al. }\\ (2021)~\cite{Shibata2021} \end{tabular}}
			& Class-level selective forgetting in lifelong learning& 
			\begin{tabitemize}[leftmargin=*,noitemsep,topsep=0pt,partopsep=0pt]
				\item Mnemonic codes
				\item New loss function with four components
			\end{tabitemize}
			& \begin{tabitemize}[leftmargin=*,noitemsep,topsep=0pt,partopsep=0pt]
				\item Class-level removing
				\item No need for original data
			\end{tabitemize}
			&
			\begin{tabitemize}[leftmargin=*,noitemsep,topsep=0pt,partopsep=0pt]
				\item Customized for image data
				\item Computational cost of embedding codes
			\end{tabitemize}
			\\ \hline
		\end{tabular}
	}
\end{table*}

\textbf{\textit{Comparisons and Discussions. } }
Recent research on selective forgetting for deep neural networks has pursued re-optimization strategies to update models and reduce the influence of data points to be removed. These works have shown promising progress while revealing key challenges and opportunities for improvement. The summary and comparison of approximate unlearning based on re-optimization are in Table~\ref{tbreopt}. 

Earlier work~\cite{golatkar2020eternal} introduces core techniques such as weight scrubbing and adding noise, providing a theoretical framework to bound residual information. However, as noted in their late work~\cite{golatkar2020forgetting}, these weight-centric analyses failed to fully capture the information remaining in activations, leading to cleaner removal guarantees by analyzing outputs. 

One challenge lies in enhancing computational efficiency. Approximations using the Fisher information matrix~\cite{golatkar2020eternal} or NTK~\cite{golatkar2020forgetting} help address scalability but may still be expensive and rely on simplifying assumptions. An interesting approach involves separating weights into fixed core and trainable user components is an interesting way. However, its dependence on strong convexity and linear approximations may limit its generalization ability. Concurrently, research~\cite{Shibata2021} used memory codes to enable class-level removing, but stability and transferability over multiple tasks still need to be proven.

Another key challenge is quantifying the closeness between original and re-optimized models, which impacts unlearning effectiveness. The more similar the re-optimized model is to one trained without the data to be removed, the less extractable information remains. However, accurately bounding this similarity is difficult for complex deep networks. Developing rigorous verification methods is an important open problem. 

In addition, these approaches make simplifying assumptions about training processes and loss landscapes that may not fully capture the intricate behaviors of large non-convex models optimized with SGD. Relaxing assumptions such as quadratic losses could improve generalization. 

Finally, inherent tradeoffs exist between privacy, accuracy, and efficiency in approximate unlearning. The isolation of user weights~\cite{golatkar2021mixed} is an interesting architectural adaptation that could enable broader applications for compartmentalizing model knowledge. Extending this approach merits further exploration. 

Overall, approximate unlearning based on re-optimization shows promising early progress on efficient data removal for deep learning. Advancing theoretical foundations, scaling to large models, and developing robust algorithms tailored to complex training dynamics could further accelerate progress in this critical area.

\begin{table*}[htbp]
	\centering
	\caption{Approximate Unlearning based on Gradient Update}\label{tbgd}
	\resizebox{\textwidth}{!}{
		\begin{tabular}{m{1.2cm}m{2.5cm}m{3.3cm}m{4cm}m{3.5cm} }
			\hline
			\textbf{Paper} & \textbf{Goal} & \textbf{Design Ideas} & \textbf{Strengths} & \textbf{Weaknesses} \\ \hline
			\begin{tabular}{@{}c@{}}DeltaGrad\\~\cite{wu2020deltagrad} \end{tabular}& Efficiently retrain models after minor data changes & 
			\begin{tabitemize}[leftmargin=*, noitemsep, topsep=0pt, partopsep=0pt] 
				\item Cached training parameters and gradients
				\item Burn-in iterations + L-BFGS approximation
			\end{tabitemize} & 
			\begin{tabitemize}[leftmargin=*, noitemsep, topsep=0pt, partopsep=0pt] 
				\item Applicable to general models with GD/SGD
				\item Support both addition and removal
			\end{tabitemize} & 
			\begin{tabitemize}[leftmargin=*, noitemsep, topsep=0pt, partopsep=0pt] 
				\item Gradient storage cost
				\item Relies on strong convexity assumptions
			\end{tabitemize} \\ \hline
			
			\begin{tabular}{@{}c@{}}FedRecover\\~\cite{cao2022fedrecover}\end{tabular}& FL model recovery after poisoning attacks& 
			\begin{tabitemize}[leftmargin=*, noitemsep, topsep=0pt, partopsep=0pt] 
				\item Server estimates updates
				\item L-BFGS approximation +corrections
			\end{tabitemize} & 
			\begin{tabitemize}[leftmargin=*, noitemsep, topsep=0pt, partopsep=0pt] 
				\item Estimate clients' model updates to reduce communication cost
				\item Scalable to numerous clients
			\end{tabitemize} & 
			\begin{tabitemize}[leftmargin=*, noitemsep, topsep=0pt, partopsep=0pt] 
				\item Storing historical information
				\item Convexity assumptions
			\end{tabitemize} \\ \hline
			
			\begin{tabular}{@{}c@{}}Descent-to\\-Delete\\~\cite{neel2021descent} \end{tabular}& Unlearning with efficiency and privacy guarantees & 
			\begin{tabitemize}[leftmargin=*, noitemsep, topsep=0pt, partopsep=0pt] 
				\item Gradient descent perturbations
				\item Data partitioning
			\end{tabitemize} & 
			\begin{tabitemize}[leftmargin=*, noitemsep, topsep=0pt, partopsep=0pt] 
				\item Gaussian noise for indistinguishability
				\item Handles arbitrary updates
				\item Improved accuracy for high-dimensional data
			\end{tabitemize} & 
			\begin{tabitemize}[leftmargin=*, noitemsep, topsep=0pt, partopsep=0pt] 
				\item Convexity assumptions
				\item Accuracy/efficiency tradeoff
			\end{tabitemize} \\ \hline	
			\begin{tabular}{@{}c@{}}BAERASER\\~\cite{liu2022backdoor} \end{tabular} & Remove backdoor& 
			\begin{tabitemize}[leftmargin=*, noitemsep, topsep=0pt, partopsep=0pt] 
				\item Trigger pattern recovery
				\item Gradient ascent unlearning
			\end{tabitemize} & 
			\begin{tabitemize}[leftmargin=*, noitemsep, topsep=0pt, partopsep=0pt] 
				\item Removes backdoors without retraining data
				\item Prevents catastrophic forgetting
			\end{tabitemize} & 
			\begin{tabitemize}[leftmargin=*, noitemsep, topsep=0pt, partopsep=0pt] 
				\item Recovered triggers not identical
				\item Applicable to backdoor only
			\end{tabitemize} \\ \hline
		\end{tabular}
	}
\end{table*}

\subsection{Approximate Unlearning based on Gradient Update}\label{secgd} 
Approximate unlearning based on gradient updates makes small adjustments to model parameters to modify the model after incrementally removing or adding data points. These methods generally follow a two-step framework to update trained models after minor data changes without full retraining:
\begin{enumerate} [label=(\arabic*)] 
	\item Initialize the model parameters using the previously trained model.
	\item Perform a few gradient update steps on the new data.
\end{enumerate}

DeltaGrad~\cite{wu2020deltagrad}, a representative of this category, adapts models efficiently to small training set changes by utilizing cached gradient and parameter information during the original training process. The algorithm includes two cases: burn-in iteration and other iterations.

\textbf{ Before Update:} The model parameters $\mathbf{w}_0, \mathbf{w}_1, ..., \mathbf{w}_t$ and corresponding gradients $\nabla F(\mathcal{D};\mathbf{w}_0),\nabla F(\mathcal{D};\mathbf{w}_1)$, $...$, $\nabla F(\mathcal{D};\mathbf{w}_t) $ of the training process on the full training dataset are cached.

\textbf{Burn-in Iteration:} The algorithm computes gradients exactly in initial burn-in iterations for correction:
\begin{equation}
	\begin{split}
		&\nabla F\left(\mathcal{D};\mathbf{w}_t^I\right)=\nabla F\left(\mathcal{D};\mathbf{w}_t\right)+\mathbf{H}_t \cdot\left(\mathbf{w}_t^I-\mathbf{w}_t\right),\\
		&\mathbf{w}_{t+1}^I=\mathbf{w}_t^I-\frac{\eta_t}{n-r}\left[\sum_{z \notin U} \nabla f\left(z;\mathbf{w}_t^I\right)\right]
	\end{split}
\end{equation}
$\mathbf{w}_{t}^I$ denotes the updated model parameter, and $\mathbf{H}_t=
\int_0^1 \mathbf{H}\left(\mathbf{w}_t+x\left(\mathbf{w}^I{ }_t-\mathbf{w}_t\right)\right) \text{d} x$ is the integrated Hessian matrix at iteration step $t$. 

\textbf{Other Iteration:} The algorithm approximates $\mathbf{H}_t$ using the L-BFGS algorithm~\cite{byrd1994representations} and uses this approximation $\mathbf{B}_{t}$ to compute updated gradients:
\begin{equation}
	\begin{split}
		&\nabla F\left(\mathcal{D};\mathbf{w}_t^I\right)=\nabla F\left(\mathcal{D};\mathbf{w}_t\right)+\mathbf{B}_t \cdot\left(\mathbf{w}_t^I-\mathbf{w}_t\right),\\
		&\mathbf{w}_{t+1}^I=\mathbf{w}_t^I-\frac{\eta_t}{n-r}\left[n \nabla F\left(\mathcal{D};\mathbf{w}_t^I\right)-\sum_{z' \in U} \nabla f\left(z';\mathbf{w}_t^I\right)\right]
	\end{split}
\end{equation}

FedRecover~\cite{cao2022fedrecover} takes a similar approach to recover accurate global models from poisoned models in federated learning while minimizing computation and communication costs on the client side. The key idea is that the server uses the historical information collected during the training of the poisoned global model to estimate the client's model update during recovery. FedRecover also utilizes  L-BFGS~\cite{byrd1994representations} to approximate the integral Hessian matrix and recover an accurate global model using strategies such as warm-up, periodic correction, and final tuning.

Descent-to-Delete~\cite{neel2021descent} introduces a basic gradient descent algorithm that begins with the previous model and executes a limited number of gradient descent updates. This process ensures the model parameters remain in close Euclidean proximity to the optimal parameters. Gaussian noise is applied to the model parameters to ensure indistinguishability for any entity close to the optimal model. For high-dimensional data, it partitions the data and independently optimizes each partition, releasing a perturbed average similar to FederatedAveraging~\cite{mcmahan2017communication}.

BAERASER~\cite{liu2022backdoor} applies gradient ascent-based unlearning to remove backdoors~\cite{zhong2020backdoor} in models. 
The process begins by identifying embedded trigger patterns. Once these triggers are discovered, BAERASER uses them to discard the contaminated memories through a gradient ascent-based machine unlearning method. 
The unlearning is designed to maximize the cross-entropy loss between the model's prediction for a trigger pattern and the target label, thereby reducing the influence of the trigger pattern. To prevent the model's performance from dropping due to the unlearning process, BAERASER uses the validation data to maintain the memory of the target model over the normal data and a dynamic penalty mechanism to punish the over-unlearning of the memorizes unrelated to trigger patterns. 

\textbf{\textit{Comparisons and Discussions. } } 
Approximate unlearning based on gradient update can use cached information such as gradients and parameters to rapidly adapt models to small data changes. Table~\ref{tbgd} summarizes and compares approximate unlearning based on gradient update. 

A key advantage is rapidly adapting models to small data changes with minimal computational expense. However, the techniques rely on assumptions such as strong convexity that may not hold for complex models~\cite{wu2020deltagrad,neel2021descent}. Techniques such as L-BFGS that approximate Hessians to speed up~\cite {wu2020deltagrad, cao2022fedrecover} may also break down for very high-dimensional models.

Another limitation is approximation errors can accumulate over multiple update rounds~\cite{neel2021descent}, resulting in less accurate recovered models. Strategies such as warm-up and periodic correction~\cite{wu2020deltagrad, cao2022fedrecover} address this but introduce extra costs. The techniques also struggle with large data changes, as the gradient adjustments are insufficient to adapt the model adequately. 

In summary, these gradient-based unlearning methods offer promising efficiency gains for data removal, but practical deployments must carefully validate assumptions and theoretical guarantees. Improvements in robustness to approximation errors and large data changes could expand their applicability.

\subsection{Approximate Graph Unlearning}\label{secgraph} 
Graph-structured data brings unique challenges to machine unlearning due to the inherent dependencies between connected data points. Traditional machine unlearning methods designed for independent data often fail to account for the complex interactions present in graph data. This section introduces several works that have made important contributions to graph-based models. These works propose methods to address the challenges of graph unlearning from different perspectives, which are crucial for a deep understanding of the current state of research in this field.

First, data interdependence is a key challenge in graph unlearning. Given a node in a graph as a removing target, it is necessary to remove its influence and its potential influence on multi-hop neighbors. To address this issue, Wu \textit{et al. }~\cite{wu2023gif} proposed a Graph Influence Function (GIF) to consider such structural influence of node/edge/feature on its neighbors. GIF estimates the parameter changes in response to $\epsilon$-mass perturbations in the removed data by introducing an additional loss term related to the affected neighbors. GIF provides a way to explain the effects of unlearning node features.
Cheng \textit{et al.}~\cite{cheng2022gnndelete} proposed GNNDELETE, a method that integrates a novel deletion operator to address the impact of edge deletion in graphs. They introduced two key properties, namely deleted edge consistency and neighborhood influence, to limit the impact of edge deletion to only the local neighborhood. Deleted edge consistency ensures that the deletion of an edge does not affect the representation of other edges in the same neighborhood. Neighborhood influence ensures that the deletion of an edge only affects its direct neighbors and not the entire graph. 
Chien~\cite{chien2022efficient} aims to address three types of data removal requests in graph unlearning: node feature unlearning, edge unlearning, and node unlearning. They derive theoretical guarantees for node/edge/feature deletion specifically for simple graph convolutions and their generalized PageRank generalizations.

Second, most graph unlearning methods are designed for the transductive graph setting, where the graph is static, and test graph information is available during training. However, many real-world graphs are dynamic, continuously adding new nodes and edges. To address this, Wang \textit{ et al.}~\cite{wang2023inductive} proposed the GUIded InDuctivE Graph Unlearning framework (GUIDE) to realize graph unlearning for dynamic graphs. GUIDE includes fair and balanced guided graph partitioning, efficient subgraph repair, and similarity-based aggregation. Balanced partitioning ensures that the retraining time of each shard is similar, and subgraph repair and similarity-based aggregation reduce the side effects of graph partitioning, thereby improving model utility.
\begin{table*}
	\centering
	\caption{Approximate Graph Unlearning Methods} \label{tabgraph}
	\resizebox{\textwidth}{!}{
		\begin{tabular}{m{1.2cm}m{2cm}m{3.5cm}m{4cm}m{3.5cm} }
			\hline
			\textbf{Paper} & \textbf{Goal} & \textbf{Design Ideas} & \textbf{Strengths} & \textbf{Weaknesses} \\ \hline
			GIF~\cite{wu2023gif} & General unlearning strategy for GNNs & Graph influence functions considering neighbors' influence & 
			\begin{tabitemize}[leftmargin=*, noitemsep, topsep=0pt, partopsep=0pt]
				\item Applicable to different models 
				\item Supports various removal tasks
				\item Closed-form solution 
			\end{tabitemize} & 
			\begin{tabitemize}[leftmargin=*, noitemsep, topsep=0pt, partopsep=0pt]
				\item Focus on the classification
				\item Memory-intensive
			\end{tabitemize} \\ \hline
			\raisebox{0.5\height}{\begin{tabular}{@{}l@{}}GNNDelete\\ ~\cite{cheng2022gnndelete}  \end{tabular}}
			& Efficient unlearning for GNNs & 
			\begin{tabitemize}[leftmargin=*, noitemsep, topsep=0pt, partopsep=0pt]
				\item Deleted Edge Consistency 
				\item Neighborhood Influence 
			\end{tabitemize} & 
			\begin{tabitemize}[leftmargin=*, noitemsep, topsep=0pt, partopsep=0pt]
				\item Flexible removal operator applicable to any GNN
				\item Supports various removal tasks
			\end{tabitemize}& 
			\begin{tabitemize}[leftmargin=*, noitemsep, topsep=0pt, partopsep=0pt]
				\item Limited to transductive learning
			\end{tabitemize} \\ \hline
			\raisebox{0.5\height}{\begin{tabular}{@{}l@{}}Chien \textit{et al. }\\ (2022)~\cite{chien2022efficient} \end{tabular}}
			& Graph-structured data unlearning& 
			\begin{tabitemize}[leftmargin=*, noitemsep, topsep=0pt, partopsep=0pt]
				\item Update model parameters based on gradient difference
			\end{tabitemize} & 
			\begin{tabitemize}[leftmargin=*, noitemsep, topsep=0pt, partopsep=0pt]
				\item Analyze for node/edge/feature unlearning
				\item Strong theoretical guarantees
			\end{tabitemize} & 
			\begin{tabitemize}[leftmargin=*, noitemsep, topsep=0pt, partopsep=0pt]
				\item Analysis limited to linear models
				\item Loose worst-case bounds
			\end{tabitemize} \\ \hline
			GUIDE\cite{wang2023inductive} &  Inductive graph unlearning & 
			\begin{tabitemize}[leftmargin=*, noitemsep, topsep=0pt, partopsep=0pt]
				\item Guided graph partitioning
				\item Subgraph repairing
				\item Similarity-based aggregation
			\end{tabitemize} & 
			\begin{tabitemize}[leftmargin=*, noitemsep, topsep=0pt, partopsep=0pt]
				\item Repair subgraphs independently
				\item Enables unlearning on evolving graphs
			\end{tabitemize} & 
			\begin{tabitemize}[leftmargin=*, noitemsep, topsep=0pt, partopsep=0pt]
				\item Increased memory cost
				\item Generalizing partition fairness needs exploration
			\end{tabitemize} \\ \hline
			\raisebox{0.5\height}{\begin{tabular}{@{}l@{}}Pan\textit{et al. }\\ (2023)~\cite{pan2023unlearning} \end{tabular}}
			& Unlearning for GNNs with limited training data & Use GSTs for graph embeddings & 
			\begin{tabitemize}[leftmargin=*, noitemsep, topsep=0pt, partopsep=0pt]
				\item Nonlinear approximate graph unlearning
				\item Theoretical guarantees
			\end{tabitemize} & 
			\begin{tabitemize}[leftmargin=*, noitemsep, topsep=0pt, partopsep=0pt]
				\item Limited to classification
				\item Bounds are loose
			\end{tabitemize} \\ \hline
			FedLU\cite{zhu2023heterogeneous} & Unlearning for knowledge graphs in FL & 
			\begin{tabitemize}[leftmargin=*, noitemsep, topsep=0pt, partopsep=0pt]
				\item Mutual knowledge distillation 
				\item Retroactive interference \& passive decay for unlearning
			\end{tabitemize} & 
			\begin{tabitemize}[leftmargin=*, noitemsep, topsep=0pt, partopsep=0pt]
				\item Unlearning for heterogeneous federated KGs
				\item Mutual knowledge distillation reduces bias in local and global models
			\end{tabitemize} & 
			\begin{tabitemize}[leftmargin=*, noitemsep, topsep=0pt, partopsep=0pt]
				\item High computational cost
				\item Applicability to other graph data needs exploration
			\end{tabitemize} 
			\\ \hline
		\end{tabular}
	}
\end{table*}

Third, it is more challenging to achieve graph unlearning while maintaining model performance when the number of training data is limited. To address this, Pan\textit{ et al.}~\cite{pan2023unlearning} proposed a nonlinear approximate graph unlearning method based on Graph Scattering Transform (GST). GST is stable under small perturbations in graph features and topologies, making it a robust method for graph data processing. Furthermore, GSTs are non-trainable, and all wavelet coefficients in GSTs are constructed analytically, making GST computationally more efficient and requiring less training data than GNNs.

Finally, in the realm of federated knowledge graph (KG) embedding learning, a significant concern is the effective handling of data heterogeneity and knowledge retention challenges. FedLU~\cite{zhu2023heterogeneous} uses mutual knowledge distillation to transfer local knowledge to the global model and absorb global knowledge into the local models. FedLU adopts a two-step forgetting process of interference and decay. In the first retroactive interference step, FedLU performs hard and soft confusions, as the interference theory states that forgetting occurs when memories compete and interfere with other memories\cite{wixted2021role}. Then, in the passive decay step, FedLU suppresses the activation of forgotten knowledge, as the decay theory posits that memory traces gradually disappear and are eventually lost if not retrieved and rehearsed~\cite{hardt2013decay}.

\textbf{\textit{Comparisons and Discussions. } }
Recent research on graph unlearning has extended machine unlearning techniques to accommodate the additional complexities of graph-structured data and node dependencies. However, there remain open challenges and opportunities for future work. The summary and comparison of approximate graph unlearning methods are in Table~\ref{tabgraph}.

Graph unlearning requires specialized techniques considering graph data's unique dependencies and constraints. Simple adaptations of existing unlearning methods are insufficient. Research~\cite{wu2023gif} proposes graph-oriented influence functions by incorporating the loss of influenced neighbors, which outperforms conventional influence functions in GNN.  

Another limitation of current graph unlearning methods is scalability to large and dynamic graphs~\cite{zhu2023heterogeneous,wang2023inductive}. Most techniques are only evaluated on small networks and may not extend well to graphs with millions of nodes and edges that evolve over time. Developing efficient and incremental unlearning algorithms is important for enabling real-world deployment.

An interesting area for further exploration is applying graph unlearning in advanced graph-based applications such as recommender systems ~\cite{pan2023unlearning,zhu2023heterogeneous}, node classification, and link prediction. Studying the influence of unlearning parts of a knowledge graph on downstream predictive tasks could provide insight into how much model utility is retained.

Overall, graph unlearning still has rich opportunities for impactful contributions through innovations in efficiency, scalability, flexibility, and rigorous characterization. By addressing the unique needs of graph data, future advances can expand the scope and applications of machine unlearning.
\subsection{Approximate Unlearning based on Novel Techniques}\label{secother}
Researchers have also developed novel approximate unlearning techniques that exploit unique model architectures or data characteristics. 

Wang \textit{et al.}~\cite{wang2022federated} propose model pruning for selectively removing categories from trained CNN classification models in FL. This approach is based on the observation that different CNN channels contribute differently to image categories. They use Term Frequency Inverse Document Frequency (TF-IDF) to quantify the class discrimination of channels and prune highly discriminative channels of target categories to facilitate unlearning. When the federated unlearning process begins, the federated server notifies clients to calculate and upload local representations. The server then prunes discriminative channels and fine-tunes the model to regain accuracy, avoiding full retraining.

Izzo \textit{et al.}~\cite{izzo2021approximate} propose the Projective Residual Update (PRU) for data removal from linear regression models. PRU computes the projection of the exact parameter update vector onto a specific low-dimensional subspace, with its computational cost scaling linearly with the data dimension, making it independent of the number of training data points. The PRU algorithm begins by creating synthetic data points and computing Leave-k-out (LKO) predictions. Next, the pseudoinverse of a matrix, composed of the outer product of the feature vectors of the removed data points, is computed. The model's loss at these synthetic points is minimized by taking a gradient step, which also updates the model parameters. This characteristic makes PRU a scalable solution for handling large datasets, as the volume of training data does not compromise its efficiency.

ERM-KTP~\cite{lin2023erm} is an interpretable knowledge-level machine unlearning method that removes target classes' influence for image classifiers. To achieve this, ERM-KTP employs an Entanglement-Reduced Mask (ERM) during training to separate and isolate class-specific knowledge. When unlearning is needed, Knowledge Transfer and Prohibition (KTP) selectively transfers non-target knowledge to a new model while prohibiting target knowledge transfer. The interpretable design enhances trust and transparency in the unlearning process.

Boundary Unlearning~\cite{chen2023boundary} aims to efficiently erase an entire class from a trained deep neural network by shifting the decision boundary instead of modifying parameters. It transfers attention from the high-dimensional parameter space to the decision space of the model, allowing rapid removal of the target class. Two methods are introduced: boundary shrink reassigns each removing data point to its nearest incorrect class and fine-tunes the model accordingly; boundary expanding temporarily maps all removing data to a new shadow class, fine-tunes the expanded model, and then prunes away the shadow class. 

Quark~\cite{lu2022quark} aims to unlearn undesirable behaviors such as toxicity and repetition from large pre-trained language models using reinforcement learning techniques. It works by scoring model samples with a reward function and sorting into quantiles, appending special reward tokens denoting the quantile, retraining the model on each quantile conditioned on its token, with a KL penalty, and sampling at generation time with the best reward token to steer towards higher rewards.

\section{Critical Issues of Machine Unlearning}\label{secdis}
\subsection{Performance and Privacy Issues of Unlearning}
Machine unlearning provides the ability to selectively remove previously learned data from trained models~\cite{cao2015towards}. As a supplementary tool to traditional ML models, machine unlearning enables efficient modification, updating, and refining of models. This subtractive capability facilitates use cases such as removing personal data for privacy, resisting poisoned data, and responding to new regulations.

However, machine unlearning also brings technical challenges and tradeoffs in ML system design. One challenge is the tradeoff of unlearning efficiency with model utility. Architectures such as SISA improve unlearning efficiency through ensemble models that isolate data but reduce model performance~\cite{ma2022learn}. Other unlearning algorithms that maintain the original ML model structure may still decrease model accuracy and even lead to catastrophic forgetting~\cite{ye2022learning,nguyen2022markov}. The unlearning methods must be tailored to minimize negative impacts on the model. Another potential issue is that removing one data point may expose information about other data points, compromising privacy~\cite{carlini2022privacy,suriyakumar2022algorithms}. It is necessary to combine machine unlearning with encryption techniques to mitigate such risks.

To tackle these challenges, advances in model architectures, dataset engineering, and infrastructure to support machine unlearning are required. For example, new model architectures could isolate data to limit negative impacts during unlearning. Synthetic dataset generation can create training data with built-in controls for later unlearning. With solutions across model, data, and system levels, machine unlearning can become a fundamental technique to construct more trustworthy, secure, and privacy-preserving ML systems.

\subsection{Machine Unlearning and the Right to Be Forgotten}
The right to be forgotten, allowing individuals to request personal data removal, is important for privacy protection. Although machine unlearning became popular partly because it enabled the right to be forgotten, the relationship between machine unlearning and this right is not a direct or exclusive binding. Machine unlearning is a useful technical tool, but it is neither necessary nor sufficient to guarantee the right to be forgotten.

First, machine unlearning is not strictly necessary for exercising the right to be forgotten, as other techniques, such as retraining models from scratch or using new datasets, can also fulfill data-removing requirements. Machine unlearning provides a more computationally efficient method for forgetting, but other feasible approaches exist. 

Second, machine unlearning alone is insufficient for comprehensively guaranteeing the right to be forgotten. Beyond removing model parameters through machine unlearning algorithms, additional technical and legal steps are required to fully assert this right in practice, such as verifiable proof of unlearning, proof of data ownership, auditing for potential privacy leaks, and employing privacy-enhancing technologies. 

Third, adapting machine unlearning for the right to be forgotten will also bring new threats. For example, malicious data owners could continuously initiate unlearning requests to decrease the model availability of model owners. Defending against such attacks remains an open challenge. Additionally, in data sharing, multiple parties share their data to train a common machine learning model. Malicious data owners could deliberately share low-quality data, unlearn it after obtaining the model, then retrain on their high-quality data to gain advantage~\cite{nguyen2020variational}.

Stronger legal and technical tools integration are needed to balance better users' rights and company interests~\cite{villaronga2018humans}. Aligning regulations and technical standards can enable compliant, minimized-cost unlearning algorithms~\cite{suriyakumar2022algorithms}. Machine unlearning provides useful techniques but must be implemented alongside other safeguards to encourage companies to act quickly to assert users' data protection rights.

\subsection{Proof of Unlearning}
Proof of unlearning is an important concept that provides auditable evidence that an ML model has properly unlearned certain training data points. While related to proof of learning~\cite{jia2021proof,zhang2022adversarial}, proof of unlearning is more challenging as the adversary is the model owner themselves, since they have access to the original training data and model~\cite{thudi2022necessity}. Stronger cryptographic proofs are needed to prevent manipulation of the audit process.

Recent work demonstrates that unlearning is best defined at the algorithmic level rather than by reasoning about model parameters~\cite{thudi2022necessity}. They show that model parameters can be identical even when trained with or without certain data points~\cite{shumailov2021manipulating}. This means inspecting parameters alone cannot prove whether unlearning occurred~\cite{thudi2022necessity}. Other related work has proposed using techniques such as backdoors to detect compliance of a model owner with a data removal request~\cite{gao2022verifi, sommer2022athena}. Such techniques are only probabilistic in their success and are constrained by the data points set as backdoors.

To enable rigorous auditing, continued research on efficient cryptographic proofs for unlearning is important. Promising directions include succinct zero-knowledge proofs and Trusted Execution Environments (TEEs)~\cite{weng2022proof}. These can prove unlearning algorithms were properly executed without relying on indirect model observations. Key challenges include ensuring proofs' transparency and reliability and minimizing verification time and cost on the underlying machine learning pipelines.

By combining cryptography, trusted computing, and advances in machine learning, proof of unlearning can provide robust auditing of unlearning claims. This emerging technique will be essential for accountability and compliance as the right to be forgotten is increasingly recognized. Ongoing research to develop practical proofs will support reliable validation of unlearning in complex, real-world systems.

\subsection{Unlearning and Explainable AI}

There is an intricate relationship between explainable AI and machine unlearning. 
On the one hand, improving model explainability can facilitate the development of more effective unlearning algorithms. Techniques from explainable AI can help identify the model components most influenced by specific training data. Such insights can guide the design of unlearning methods that precisely remove the targeted information. Influence function-based unlearning is a prime example of explainable AI solving unlearning issues. By quantifying the influence of each training sample on model parameters, it allows selective removal of influential data points. However, as discussed in Section \ref{secif}, influence function estimation remains computationally expensive and numerically unstable. Developing more efficient approximations of influence is an active research area.

Machine unlearning also contributes to the advancement of explainable AI. For example, we can gain insights into how the removed information influenced the model by analyzing how explanations of model predictions evolve before and after unlearning specific data. Comparing explanations derived from the original model versus the updated model after unlearning can elucidate what memorization and unlearning entail in complex machine learning systems~\cite{lundberg2017unified}. 

Overall, explainability and unlearning are mutually beneficial, and integrating the two leads to more transparent, trustworthy, and controllable AI systems. There remain open challenges around developing quantitative evaluation protocols to assess the interplay between explainability and unlearning.

\section{Potential Research Directions}\label{secpotential}
Machine unlearning is an emerging field with many open problems and opportunities. This section discusses several promising directions for advancing machine unlearning research.  

\subsection{Unlearning for Diverse Data Structures}
Existing studies have focused on exploring unlearning algorithms on set-structured and graph-structured training data. However, many other complex data structures such as text, speech, and multimedia are also increasingly used to train models, especially in fields such as natural language processing~\cite{hirschberg2015advances} and audio/video analysis~\cite{takahashi2017aenet}. Extending unlearning techniques to these diverse and complex data is an important next step but poses challenges. This requires handling the unique characteristics of different data types, such as timing and sequences in speech and language, spatial relationships in images, or hierarchical structures in knowledge graphs. Further, developing multimodal unlearning~\cite{ngiam2011multimodal} that works across data combinations is also crucial for real-world usage. Tackling these data structure challenges can greatly expand the applicability of machine unlearning.

\subsection{Unlearning for Non-convex Models}
Non-convex models, such as CNNs, Recurrent Neural Networks (RNNs), and Transformers, have gained widespread adoption across various domains. However, existing research on approximate unlearning has predominantly focused on convex models. Extending efficient and effective unlearning algorithms to non-convex neural networks remains an important open challenge ~\cite{guo2020certified}. The key difficulties include dealing with non-convex optimization problems such as local optima and saddle points and providing theoretical guarantees on the unlearning process similar to those for convex cases. Additionally, adapting to the unique structures in neural networks, such as activation functions and normalization layers~\cite{ioffe2015batch}, poses further complexities.

\subsection{User-Specified Granularity of Unlearning }
Most existing machine unlearning methods focus on instance-level removing, i.e., removing the influence of one training data point. However, users may need finer-grained control over what to remove from the model. For example, users may request to remove only certain sensitive areas of an image while retaining the rest or specific words in a text document that are no longer appropriate. An interesting research direction is to explore interactive and interpretable unlearning algorithms that allow users to specify the granularity of unlearning at a finer-grained level. Such algorithms need to identify the semantic components of examples and their contributions to model predictions. It can greatly enhance the ability of unlearning techniques to meet user requirements. 

\subsection{Privacy Assurance for Unlearning}
Most existing approximate unlearning algorithms rely on differential privacy to provide formal unlearning guarantees ~\cite{guo2020certified,gupta2021adaptive}. However, differential privacy often uses a relatively relaxed privacy budget to balance privacy and utility~\cite{jagielski2020auditing}. Its privacy guarantees may be insufficient for scenarios with extremely high privacy demands~\cite{jayaraman2019evaluating,hu2015differential}.
Therefore, an important research direction is exploring stronger notions beyond differential privacy that can limit information disclosure rigorously while not excessively sacrificing model utility. For example, exploring information theoretic approaches~\cite{shannon1948mathematical} that directly bound the amount of information about the removed data retained in the model after unlearning. This requires overcoming the difficulty of extracting information from models. This research direction can potentially promote the explainability and verifiability of machine learning algorithms~\cite{eisenhofer2022verifiable}.

\subsection{Quantitative Evaluation Metrics}
To compare different unlearning methods, it is crucial to develop quantitative metrics that can measure the degree of influence removal for the removed data and the degree of influence retention for the remaining data~\cite{weng2022proof,thudi2022unrolling}. However, constructing such fine-grained evaluation metrics requires the ability to systematically analyze the memorization process of machine learning models on different data~\cite{thudi2022unrolling}. Advanced tools from information theory~\cite{tishby2000information} and explainability research~\cite{molnar2020interpretable} are useful in this direction. Well-designed metrics can greatly promote the theoretical analysis of machine unlearning algorithms and guide the development and adoption of reliable unlearning techniques in practice.

\section{Conclusion} \label{seccon}
This paper presents a comprehensive overview of machine unlearning, an emerging technique that enables selective removing training data in machine learning models. We reviewed exact unlearning solutions and approximate unlearning solutions based on influence functions, re-optimization, gradient update, and other approaches. Our analysis reveals current limitations in computational complexity, providing strict privacy guarantees and limiting model utility. We suggest promising directions such as extending unlearning to diverse models and data structures, developing efficient and verifiable algorithms, establishing rigorous evaluation methods, and exploring connections with other related fields. By addressing these challenges from different aspects, such as algorithms, systems, and regulations, machine unlearning can become an integral capability for trustworthy and adaptive AI.

\bibliographystyle{IEEEtran}
\bibliography{cite}

\begin{thebibliography}{100}
\providecommand{\url}[1]{#1}
\csname url@samestyle\endcsname
\providecommand{\newblock}{\relax}
\providecommand{\bibinfo}[2]{#2}
\providecommand{\BIBentrySTDinterwordspacing}{\spaceskip=0pt\relax}
\providecommand{\BIBentryALTinterwordstretchfactor}{4}
\providecommand{\BIBentryALTinterwordspacing}{\spaceskip=\fontdimen2\font plus
\BIBentryALTinterwordstretchfactor\fontdimen3\font minus
  \fontdimen4\font\relax}
\providecommand{\BIBforeignlanguage}[2]{{%
\expandafter\ifx\csname l@#1\endcsname\relax
\typeout{** WARNING: IEEEtran.bst: No hyphenation pattern has been}%
\typeout{** loaded for the language `#1'. Using the pattern for}%
\typeout{** the default language instead.}%
\else
\language=\csname l@#1\endcsname
\fi
#2}}
\providecommand{\BIBdecl}{\relax}
\BIBdecl

\bibitem{zhao2020exploring}
H.~Zhao, J.~Jia, and V.~Koltun, ``Exploring self-attention for image
  recognition,'' in \emph{Proceedings of the IEEE/CVF conference on computer
  vision and pattern recognition}, 2020, pp. 10\,076--10\,085.

\bibitem{hirschberg2015advances}
J.~Hirschberg and C.~D. Manning, ``Advances in natural language processing,''
  \emph{Science}, vol. 349, no. 6245, pp. 261--266, 2015.

\bibitem{wu2022graph}
S.~Wu, F.~Sun, W.~Zhang, X.~Xie, and B.~Cui, ``Graph neural networks in
  recommender systems: a survey,'' \emph{ACM Computing Surveys}, vol.~55,
  no.~5, pp. 1--37, 2022.

\bibitem{jordan2015machine}
M.~I. Jordan and T.~M. Mitchell, ``Machine learning: Trends, perspectives, and
  prospects,'' \emph{Science}, vol. 349, no. 6245, pp. 255--260, 2015.

\bibitem{fu2021knowledge}
S.~Fu, F.~He, and D.~Tao, ``Knowledge removal in sampling-based bayesian
  inference,'' in \emph{International Conference on Learning Representations},
  2021.

\bibitem{tarun2023fast}
A.~K. Tarun, V.~S. Chundawat, M.~Mandal, and M.~Kankanhalli, ``Fast yet
  effective machine unlearning,'' \emph{IEEE Transactions on Neural Networks
  and Learning Systems}, 2023.

\bibitem{jegorova2022survey}
M.~Jegorova, C.~Kaul, C.~Mayor, A.~Q. O'Neil, A.~Weir, R.~Murray-Smith, and
  S.~A. Tsaftaris, ``Survey: Leakage and privacy at inference time,''
  \emph{IEEE Transactions on Pattern Analysis and Machine Intelligence}, 2022.

\bibitem{xu2023machine}
H.~Xu, T.~Zhu*, L.~Zhang, W.~Zhou, and P.~S. Yu, ``Machine unlearning: A
  survey,'' \emph{ACM Computing Surveys}, 2023.

\bibitem{Voigt2017}
P.~Voigt and A.~Von~dem Bussche, \emph{The EU General Data Protection
  Regulation (GDPR). A Practical Guide}.\hskip 1em plus 0.5em minus 0.4em\relax
  Springer International Publishing, 2017.

\bibitem{CCPA2023}
S.~o. C. D. o.~J. Office of~the Attorney~General, ``California consumer privacy
  act (ccpa),'' \url{https://oag.ca.gov/privacy/ccpa}, 2023.

\bibitem{wang2019measure}
G.~Wang, C.~X. Dang, and Z.~Zhou, ``Measure contribution of participants in
  federated learning,'' in \emph{2019 IEEE international conference on big data
  (Big Data)}.\hskip 1em plus 0.5em minus 0.4em\relax IEEE, 2019, pp.
  2597--2604.

\bibitem{chang2021privacy}
H.~Chang and R.~Shokri, ``On the privacy risks of algorithmic fairness,'' in
  \emph{2021 IEEE European Symposium on Security and Privacy (EuroS\&P)}.\hskip
  1em plus 0.5em minus 0.4em\relax IEEE, 2021, pp. 292--303.

\bibitem{biggio2012poisoning}
B.~Biggio, B.~Nelson, P.~Laskov \emph{et~al.}, ``Poisoning attacks against
  support vector machines,'' in \emph{Proceedings of the 29th International
  Conference on Machine Learning, ICML 2012}.\hskip 1em plus 0.5em minus
  0.4em\relax ArXiv e-prints, 2012, pp. 1807--1814.

\bibitem{steinhardt2017certified}
J.~Steinhardt, P.~W.~W. Koh, and P.~S. Liang, ``Certified defenses for data
  poisoning attacks,'' \emph{Advances in neural information processing
  systems}, vol.~30, 2017.

\bibitem{gama2014survey}
J.~Gama, I.~{\v{Z}}liobait{\.e}, A.~Bifet, M.~Pechenizkiy, and A.~Bouchachia,
  ``A survey on concept drift adaptation,'' \emph{ACM computing surveys
  (CSUR)}, vol.~46, no.~4, pp. 1--37, 2014.

\bibitem{cao2015towards}
Y.~Cao and J.~Yang, ``Towards making systems forget with machine unlearning,''
  in \emph{2015 IEEE symposium on security and privacy}.\hskip 1em plus 0.5em
  minus 0.4em\relax IEEE, 2015, pp. 463--480.

\bibitem{thudi2022necessity}
A.~Thudi, H.~Jia, I.~Shumailov, and N.~Papernot, ``On the necessity of
  auditable algorithmic definitions for machine unlearning,'' in \emph{31st
  USENIX Security Symposium (USENIX Security 22)}, 2022, pp. 4007--4022.

\bibitem{Yan2022ARCANE}
H.~Yan, X.~Li, Z.~Guo, H.~Li, F.~Li, and X.~Lin, ``{ARCANE: An Efficient
  Architecture for Exact Machine Unlearning},'' in \emph{IJCAI International
  Joint Conference on Artificial Intelligence}, 2022, pp. 4006--4013.

\bibitem{doshi2017towards}
F.~Doshi-Velez and B.~Kim, ``Towards a rigorous science of interpretable
  machine learning,'' \emph{arXiv preprint arXiv:1702.08608}, 2017.

\bibitem{arrieta2020explainable}
A.~B. Arrieta, N.~D{\'\i}az-Rodr{\'\i}guez, J.~Del~Ser, A.~Bennetot, S.~Tabik,
  A.~Barbado, S.~Garc{\'\i}a, S.~Gil-L{\'o}pez, D.~Molina, R.~Benjamins
  \emph{et~al.}, ``Explainable artificial intelligence (xai): Concepts,
  taxonomies, opportunities and challenges toward responsible ai,''
  \emph{Information fusion}, vol.~58, pp. 82--115, 2020.

\bibitem{Koh2017Understanding}
P.~W. Koh and P.~Liang, ``Understanding black-box predictions via influence
  functions,'' in \emph{International conference on machine learning}.\hskip
  1em plus 0.5em minus 0.4em\relax PMLR, 2017, pp. 1885--1894.

\bibitem{hampel1974influence}
F.~R. Hampel, ``The influence curve and its role in robust estimation,''
  \emph{Journal of the american statistical association}, vol.~69, no. 346, pp.
  383--393, 1974.

\bibitem{zhou2012ensemble}
Z.-H. Zhou, \emph{Ensemble methods: foundations and algorithms}.\hskip 1em plus
  0.5em minus 0.4em\relax CRC press, 2012.

\bibitem{kuncheva2003measures}
L.~I. Kuncheva and C.~J. Whitaker, ``Measures of diversity in classifier
  ensembles and their relationship with the ensemble accuracy,'' \emph{Machine
  learning}, vol.~51, pp. 181--207, 2003.

\bibitem{rokach2010ensemble}
L.~Rokach, ``Ensemble-based classifiers,'' \emph{Artificial intelligence
  review}, vol.~33, pp. 1--39, 2010.

\bibitem{chen2021machine}
M.~Chen, Z.~Zhang, T.~Wang, M.~Backes, M.~Humbert, and Y.~Zhang, ``When machine
  unlearning jeopardizes privacy,'' in \emph{Proceedings of the 2021 ACM SIGSAC
  conference on computer and communications security}, 2021, pp. 896--911.

\bibitem{garg2020formalizing}
S.~Garg, S.~Goldwasser, and P.~N. Vasudevan, ``Formalizing data deletion in the
  context of the right to be forgotten,'' in \emph{Annual International
  Conference on the Theory and Applications of Cryptographic Techniques}.\hskip
  1em plus 0.5em minus 0.4em\relax Springer, 2020, pp. 373--402.

\bibitem{Bertram2019}
T.~Bertram, E.~Bursztein, S.~Caro, H.~Chao, R.~C. Feman, P.~Fleischer,
  A.~Gustafsson, J.~Hemerly, C.~Hibbert, L.~Invernizzi, L.~K. Donnelly,
  J.~Ketover, J.~Laefer, P.~Nicholas, Y.~Niu, H.~Obhi, D.~Price, A.~Strait,
  K.~Thomas, and A.~Verney, ``{Five years of the right to be forgotten},'' in
  \emph{Proceedings of the ACM Conference on Computer and Communications
  Security}, 2019, pp. 959--972.

\bibitem{chundawat2023zero}
V.~S. Chundawat, A.~K. Tarun, M.~Mandal, and M.~Kankanhalli, ``Zero-shot
  machine unlearning,'' \emph{IEEE Transactions on Information Forensics and
  Security}, 2023.

\bibitem{cao2018efficient}
Y.~Cao, A.~F. Yu, A.~Aday, E.~Stahl, J.~Merwine, and J.~Yang, ``Efficient
  repair of polluted machine learning systems via causal unlearning,'' in
  \emph{Proceedings of the 2018 on Asia conference on computer and
  communications security}, 2018, pp. 735--747.

\bibitem{shan2022poison}
S.~Shan, A.~N. Bhagoji, H.~Zheng, and B.~Y. Zhao, ``Poison forensics: Traceback
  of data poisoning attacks in neural networks,'' in \emph{31st USENIX Security
  Symposium (USENIX Security 22)}, 2022, pp. 3575--3592.

\bibitem{Mirzasoleiman2017}
B.~Mirzasoleiman, A.~Karbasi, and A.~Krause, ``{Deletion-robust submodular
  maximization: Data summarization with "the right to be forgotten"},'' in
  \emph{Proceddings of the 34th International Conference on Machine Learning,
  ICML 2017}, vol.~5, 2017, pp. 3780--3790.

\bibitem{serra2018overcoming}
J.~Serra, D.~Suris, M.~Miron, and A.~Karatzoglou, ``Overcoming catastrophic
  forgetting with hard attention to the task,'' in \emph{International
  conference on machine learning}.\hskip 1em plus 0.5em minus 0.4em\relax PMLR,
  2018, pp. 4548--4557.

\bibitem{Schelter2021HedgeCut}
S.~Schelter, S.~Grafberger, and T.~Dunning, ``{HedgeCut: Maintaining Randomised
  Trees for Low-Latency Machine Unlearning},'' in \emph{Proceedings of the ACM
  SIGMOD International Conference on Management of Data}, 2021, pp. 1545--1557.

\bibitem{lin2023erm}
S.~Lin, X.~Zhang, C.~Chen, X.~Chen, and W.~Susilo, ``Erm-ktp: Knowledge-level
  machine unlearning via knowledge transfer,'' in \emph{Proceedings of the
  IEEE/CVF Conference on Computer Vision and Pattern Recognition}, 2023, pp.
  20\,147--20\,155.

\bibitem{bourtoule2021machine}
L.~Bourtoule, V.~Chandrasekaran, C.~A. Choquette-Choo, H.~Jia, A.~Travers,
  B.~Zhang, D.~Lie, and N.~Papernot, ``Machine unlearning,'' in \emph{2021 IEEE
  Symposium on Security and Privacy (SP)}.\hskip 1em plus 0.5em minus
  0.4em\relax IEEE, 2021, pp. 141--159.

\bibitem{carlini2022privacy}
N.~Carlini, M.~Jagielski, C.~Zhang, N.~Papernot, A.~Terzis, and F.~Tramer,
  ``The privacy onion effect: Memorization is relative,'' \emph{Advances in
  Neural Information Processing Systems}, vol.~35, pp. 13\,263--13\,276, 2022.

\bibitem{ye2022learning}
J.~Ye, Y.~Fu, J.~Song, X.~Yang, S.~Liu, X.~Jin, M.~Song, and X.~Wang,
  ``Learning with recoverable forgetting,'' in \emph{European Conference on
  Computer Vision}.\hskip 1em plus 0.5em minus 0.4em\relax Springer, 2022, pp.
  87--103.

\bibitem{jia2021zest}
H.~Jia, H.~Chen, J.~Guan, A.~S. Shamsabadi, and N.~Papernot, ``A zest of lime:
  Towards architecture-independent model distances,'' in \emph{International
  Conference on Learning Representations}, 2021.

\bibitem{wu2020deltagrad}
Y.~Wu, E.~Dobriban, and S.~Davidson, ``{Deltagrad: Rapid retraining of machine
  learning models},'' in \emph{International Conference on Machine
  Learning}.\hskip 1em plus 0.5em minus 0.4em\relax PMLR, 2020, pp.
  10\,355--10\,366.

\bibitem{golatkar2020eternal}
A.~Golatkar, A.~Achille, and S.~Soatto, ``{Eternal sunshine of the spotless
  net: Selective forgetting in deep networks},'' in \emph{Proceedings of the
  IEEE/CVF Conference on Computer Vision and Pattern Recognition}, 2020, pp.
  9304--9312.

\bibitem{mercuri2022introduction}
S.~Mercuri, R.~Khraishi, R.~Okhrati, D.~Batra, C.~Hamill, T.~Ghasempour, and
  A.~Nowlan, ``An introduction to machine unlearning,'' \emph{arXiv preprint
  arXiv:2209.00939}, 2022.

\bibitem{guo2020certified}
C.~Guo, T.~Goldstein, A.~Hannun, and L.~Van Der~Maaten, ``Certified data
  removal from machine learning models,'' in \emph{International Conference on
  Machine Learning}.\hskip 1em plus 0.5em minus 0.4em\relax PMLR, 2020, pp.
  3832--3842.

\bibitem{dwork2006differential}
C.~Dwork, ``Differential privacy,'' in \emph{Automata, Languages and
  Programming: 33rd International Colloquium, ICALP 2006, Venice, Italy, July
  10-14, 2006, Proceedings, Part II 33}.\hskip 1em plus 0.5em minus 0.4em\relax
  Springer, 2006, pp. 1--12.

\bibitem{lecun1998mnist}
Y.~LeCun, L.~Bottou, Y.~Bengio, and P.~Haffner, ``Gradient-based learning
  applied to document recognition,'' \emph{Proceedings of the IEEE}, vol.~86,
  no.~11, pp. 2278--2324, 1998.

\bibitem{krizhevsky2009learning}
A.~Krizhevsky and G.~Hinton, ``Learning multiple layers of features from tiny
  images,'' 2009.

\bibitem{netzer2011reading}
Y.~Netzer, T.~Wang, A.~Coates, A.~Bissacco, B.~Wu, and A.~Y. Ng, ``Reading
  digits in natural images with unsupervised feature learning,'' in \emph{NIPS
  workshop on deep learning and unsupervised feature learning}, vol. 2011,
  no.~2, 2011, p.~5.

\bibitem{deng2009imagenet}
J.~Deng, W.~Dong, R.~Socher, L.-J. Li, K.~Li, and L.~Fei-Fei, ``Imagenet: A
  large-scale hierarchical image database,'' pp. 248--255, 2009.

\bibitem{maas2011learning}
A.~L. Maas, R.~E. Daly, P.~T. Pham, D.~Huang, A.~Y. Ng, and C.~Potts,
  ``Learning word vectors for sentiment analysis,'' in \emph{Proceedings of the
  49th Annual Meeting of the Association for Computational Linguistics: Human
  Language Technologies}, 2011, pp. 142--150.

\bibitem{merity2016pointer}
S.~Merity, C.~Xiong, J.~Bradbury, and R.~Socher, ``Pointer sentinel mixture
  models,'' \emph{arXiv preprint arXiv:1609.07843}, 2016.

\bibitem{Gokaslan2019OpenWeb}
A.~Gokaslan and V.~Cohen, ``Openwebtext corpus,''
  \url{http://Skylion007.github.io/OpenWebTextCorpus}, 2019.

\bibitem{sen2008collective}
P.~Sen, G.~Namata, M.~Bilgic, L.~Getoor, B.~Galligher, and T.~Eliassi-Rad,
  ``Collective classification in network data,'' \emph{AI magazine}, vol.~29,
  no.~3, pp. 93--93, 2008.

\bibitem{giles1998citeseer}
C.~L. Giles, K.~D. Bollacker, and S.~Lawrence, ``Citeseer: An automatic
  citation indexing system,'' \emph{Proceedings of the third ACM conference on
  Digital libraries}, pp. 89--98, 1998.

\bibitem{pubmed}
``Pubmed: a free search engine accessing primarily the medline database,''
  available online at: https://pubmed.ncbi.nlm.nih.gov/.

\bibitem{nguyen2022survey}
T.~T. Nguyen, T.~T. Huynh, P.~L. Nguyen, A.~W.-C. Liew, H.~Yin, and Q.~V.~H.
  Nguyen, ``A survey of machine unlearning,'' \emph{arXiv preprint
  arXiv:2209.02299}, 2022.

\bibitem{qu2023learn}
Y.~Qu, X.~Yuan, M.~Ding, W.~Ni, T.~Rakotoarivelo, and D.~Smith, ``Learn to
  unlearn: A survey on machine unlearning,'' \emph{arXiv preprint
  arXiv:2305.07512}, 2023.

\bibitem{shaik2023exploring}
T.~Shaik, X.~Tao, H.~Xie, L.~Li, X.~Zhu, and Q.~Li, ``Exploring the landscape
  of machine unlearning: A survey and taxonomy,'' \emph{arXiv preprint
  arXiv:2305.06360}, 2023.

\bibitem{Brophy2020DaRE}
\BIBentryALTinterwordspacing
J.~Brophy and D.~Lowd, ``{Machine Unlearning for Random Forests},'' in
  \emph{International Conference on Machine Learning}, sep 2020, pp.
  1092--1104. [Online]. Available:
  \url{https://proceedings.mlr.press/v139/brophy21a.html}
\BIBentrySTDinterwordspacing

\bibitem{Chen2022GraphEraser}
M.~Chen, Z.~Zhang, T.~Wang, M.~Backes, M.~Humbert, and Y.~Zhang, ``{Graph
  Unlearning},'' in \emph{Proceedings of the ACM Conference on Computer and
  Communications Security}, 2022, pp. 499--513.

\bibitem{Chen2022RecEraser}
C.~Chen, F.~Sun, M.~Zhang, and B.~Ding, ``{Recommendation Unlearning},'' in
  \emph{WWW 2022 - Proceedings of the ACM Web Conference 2022}, 2022, pp.
  2768--2777.

\bibitem{Ginart2019kmeans}
A.~A. Ginart, M.~Y. Guan, G.~Valiant, and J.~Zou, ``{Making AI forget you: Data
  deletion in machine learning},'' in \emph{Advances in Neural Information
  Processing Systems}, vol.~32, no. NeurIPS, 2019, pp. 1--14.

\bibitem{su2023asynchronous}
N.~Su and B.~Li, ``Asynchronous federated unlearning,'' in \emph{IEEE INFOCOM
  2023-IEEE Conference on Computer Communications}.\hskip 1em plus 0.5em minus
  0.4em\relax IEEE, 2023, pp. 1--10.

\bibitem{liu2022right}
Y.~Liu, L.~Xu, X.~Yuan, C.~Wang, and B.~Li, ``The right to be forgotten in
  federated learning: An efficient realization with rapid retraining,'' in
  \emph{IEEE INFOCOM 2022-IEEE Conference on Computer Communications}.\hskip
  1em plus 0.5em minus 0.4em\relax IEEE, 2022, pp. 1749--1758.

\bibitem{sekhari2021remember}
A.~Sekhari, J.~Acharya, G.~Kamath, and A.~T. Suresh, ``Remember what you want
  to forget: Algorithms for machine unlearning,'' \emph{Advances in Neural
  Information Processing Systems}, vol.~34, pp. 18\,075--18\,086, 2021.

\bibitem{suriyakumar2022algorithms}
V.~Suriyakumar and A.~C. Wilson, ``Algorithms that approximate data removal:
  New results and limitations,'' \emph{Advances in Neural Information
  Processing Systems}, vol.~35, pp. 18\,892--18\,903, 2022.

\bibitem{mehta2022deep}
R.~Mehta, S.~Pal, V.~Singh, and S.~N. Ravi, ``Deep unlearning via randomized
  conditionally independent hessians,'' in \emph{Proceedings of the IEEE/CVF
  Conference on Computer Vision and Pattern Recognition}, 2022, pp.
  10\,422--10\,431.

\bibitem{wu2022puma}
G.~Wu, M.~Hashemi, and C.~Srinivasa, ``Puma: Performance unchanged model
  augmentation for training data removal,'' in \emph{Proceedings of the AAAI
  Conference on Artificial Intelligence}, vol.~36, no.~8, 2022, pp. 8675--8682.

\bibitem{tanno2022repairing}
R.~Tanno, M.~F~Pradier, A.~Nori, and Y.~Li, ``Repairing neural networks by
  leaving the right past behind,'' \emph{Advances in Neural Information
  Processing Systems}, vol.~35, pp. 13\,132--13\,145, 2022.

\bibitem{chaudhuri2011differentially}
K.~Chaudhuri, C.~Monteleoni, and A.~D. Sarwate, ``Differentially private
  empirical risk minimization.'' \emph{Journal of Machine Learning Research},
  vol.~12, no.~3, 2011.

\bibitem{warnecke2023machine}
\BIBentryALTinterwordspacing
A.~Warnecke, L.~Pirch, C.~Wressnegger, and K.~Rieck, ``Machine unlearning of
  features and labels,'' in \emph{30th Annual Network and Distributed System
  Security Symposium, {NDSS} 2023, San Diego, California, USA, February 27 -
  March 3, 2023}.\hskip 1em plus 0.5em minus 0.4em\relax The Internet Society,
  2023. [Online]. Available:
  \url{https://www.ndss-symposium.org/ndss-paper/machine-unlearning-of-features-and-labels/}
\BIBentrySTDinterwordspacing

\bibitem{martens2020new}
J.~Martens, ``New insights and perspectives on the natural gradient method,''
  \emph{The Journal of Machine Learning Research}, vol.~21, no.~1, pp.
  5776--5851, 2020.

\bibitem{golatkar2020forgetting}
A.~Golatkar, A.~Achille, and S.~Soatto, ``Forgetting outside the box: Scrubbing
  deep networks of information accessible from input-output observations,'' in
  \emph{Computer Vision--ECCV 2020: 16th European Conference, Glasgow, UK,
  August 23--28, 2020, Proceedings, Part XXIX 16}.\hskip 1em plus 0.5em minus
  0.4em\relax Springer, 2020, pp. 383--398.

\bibitem{golatkar2021mixed}
A.~Golatkar, A.~Achille, A.~Ravichandran, M.~Polito, and S.~Soatto,
  ``Mixed-privacy forgetting in deep networks,'' in \emph{Proceedings of the
  IEEE/CVF conference on computer vision and pattern recognition}, 2021, pp.
  792--801.

\bibitem{Shibata2021}
T.~Shibata, G.~Irie, D.~Ikami, and Y.~Mitsuzumi, ``{Learning with Selective
  Forgetting},'' in \emph{IJCAI International Joint Conference on Artificial
  Intelligence}, 2021, pp. 989--996.

\bibitem{cao2022fedrecover}
X.~Cao, J.~Jia, Z.~Zhang, and N.~Z. Gong, ``Fedrecover: Recovering from
  poisoning attacks in federated learning using historical information,'' in
  \emph{2023 IEEE Symposium on Security and Privacy (SP)}.\hskip 1em plus 0.5em
  minus 0.4em\relax IEEE Computer Society, 2022, pp. 326--343.

\bibitem{neel2021descent}
S.~Neel, A.~Roth, and S.~Sharifi-Malvajerdi, ``Descent-to-delete:
  Gradient-based methods for machine unlearning,'' in \emph{Algorithmic
  Learning Theory}.\hskip 1em plus 0.5em minus 0.4em\relax PMLR, 2021, pp.
  931--962.

\bibitem{liu2022backdoor}
Y.~Liu, M.~Fan, C.~Chen, X.~Liu, Z.~Ma, L.~Wang, and J.~Ma, ``Backdoor defense
  with machine unlearning,'' in \emph{IEEE INFOCOM 2022-IEEE Conference on
  Computer Communications}.\hskip 1em plus 0.5em minus 0.4em\relax IEEE, 2022,
  pp. 280--289.

\bibitem{byrd1994representations}
R.~H. Byrd, J.~Nocedal, and R.~B. Schnabel, ``Representations of quasi-newton
  matrices and their use in limited memory methods,'' \emph{Mathematical
  Programming}, vol.~63, no. 1-3, pp. 129--156, 1994.

\bibitem{mcmahan2017communication}
B.~McMahan, E.~Moore, D.~Ramage, S.~Hampson, and B.~A. y~Arcas,
  ``Communication-efficient learning of deep networks from decentralized
  data,'' in \emph{Artificial intelligence and statistics}.\hskip 1em plus
  0.5em minus 0.4em\relax PMLR, 2017, pp. 1273--1282.

\bibitem{zhong2020backdoor}
H.~Zhong, C.~Liao, A.~C. Squicciarini, S.~Zhu, and D.~Miller, ``Backdoor
  embedding in convolutional neural network models via invisible
  perturbation,'' in \emph{Proceedings of the Tenth ACM Conference on Data and
  Application Security and Privacy}, 2020, pp. 97--108.

\bibitem{wu2023gif}
J.~Wu, Y.~Yang, Y.~Qian, Y.~Sui, X.~Wang, and X.~He, ``Gif: A general graph
  unlearning strategy via influence function,'' in \emph{Proceedings of the ACM
  Web Conference 2023}, 2023, pp. 651--661.

\bibitem{cheng2022gnndelete}
J.~Cheng, G.~Dasoulas, H.~He, C.~Agarwal, and M.~Zitnik, ``Gnndelete: A general
  strategy for unlearning in graph neural networks,'' in \emph{The Eleventh
  International Conference on Learning Representations}, 2022.

\bibitem{chien2022efficient}
E.~Chien, C.~Pan, and O.~Milenkovic, ``Efficient model updates for approximate
  unlearning of graph-structured data,'' in \emph{The Eleventh International
  Conference on Learning Representations}, 2022.

\bibitem{wang2023inductive}
C.-L. Wang, M.~Huai, and D.~Wang, ``Inductive graph unlearning,'' \emph{arXiv
  preprint arXiv:2304.03093}, 2023.

\bibitem{pan2023unlearning}
C.~Pan, E.~Chien, and O.~Milenkovic, ``Unlearning graph classifiers with
  limited data resources,'' in \emph{Proceedings of the ACM Web Conference
  2023}, 2023, pp. 716--726.

\bibitem{zhu2023heterogeneous}
X.~Zhu, G.~Li, and W.~Hu, ``Heterogeneous federated knowledge graph embedding
  learning and unlearning,'' in \emph{Proceedings of the ACM Web Conference
  2023}, 2023, pp. 2444--2454.

\bibitem{wixted2021role}
J.~T. Wixted, ``The role of retroactive interference and consolidation in
  everyday forgetting,'' in \emph{Current Issues in Memory}.\hskip 1em plus
  0.5em minus 0.4em\relax Routledge, 2021, pp. 117--143.

\bibitem{hardt2013decay}
O.~Hardt, K.~Nader, and L.~Nadel, ``Decay happens: the role of active
  forgetting in memory,'' \emph{Trends in cognitive sciences}, vol.~17, no.~3,
  pp. 111--120, 2013.

\bibitem{wang2022federated}
J.~Wang, S.~Guo, X.~Xie, and H.~Qi, ``Federated unlearning via
  class-discriminative pruning,'' in \emph{Proceedings of the ACM Web
  Conference 2022}, 2022, pp. 622--632.

\bibitem{izzo2021approximate}
Z.~Izzo, M.~A. Smart, K.~Chaudhuri, and J.~Zou, ``Approximate data deletion
  from machine learning models,'' in \emph{International Conference on
  Artificial Intelligence and Statistics}.\hskip 1em plus 0.5em minus
  0.4em\relax PMLR, 2021, pp. 2008--2016.

\bibitem{chen2023boundary}
M.~Chen, W.~Gao, G.~Liu, K.~Peng, and C.~Wang, ``Boundary unlearning: Rapid
  forgetting of deep networks via shifting the decision boundary,'' in
  \emph{Proceedings of the IEEE/CVF Conference on Computer Vision and Pattern
  Recognition}, 2023, pp. 7766--7775.

\bibitem{lu2022quark}
X.~Lu, S.~Welleck, J.~Hessel, L.~Jiang, L.~Qin, P.~West, P.~Ammanabrolu, and
  Y.~Choi, ``Quark: Controllable text generation with reinforced unlearning,''
  \emph{Advances in neural information processing systems}, vol.~35, pp.
  27\,591--27\,609, 2022.

\bibitem{ma2022learn}
Z.~Ma, Y.~Liu, X.~Liu, J.~Liu, J.~Ma, and K.~Ren, ``Learn to forget: Machine
  unlearning via neuron masking,'' \emph{IEEE Transactions on Dependable and
  Secure Computing}, 2022.

\bibitem{nguyen2022markov}
Q.~P. Nguyen, R.~Oikawa, D.~M. Divakaran, M.~C. Chan, and B.~K.~H. Low,
  ``Markov chain monte carlo-based machine unlearning: Unlearning what needs to
  be forgotten,'' in \emph{Proceedings of the 2022 ACM on Asia Conference on
  Computer and Communications Security}, 2022, pp. 351--363.

\bibitem{nguyen2020variational}
Q.~P. Nguyen, B.~K.~H. Low, and P.~Jaillet, ``Variational bayesian
  unlearning,'' \emph{Advances in Neural Information Processing Systems},
  vol.~33, pp. 16\,025--16\,036, 2020.

\bibitem{villaronga2018humans}
E.~F. Villaronga, P.~Kieseberg, and T.~Li, ``Humans forget, machines remember:
  Artificial intelligence and the right to be forgotten,'' \emph{Computer Law
  \& Security Review}, vol.~34, no.~2, pp. 304--313, 2018.

\bibitem{jia2021proof}
H.~Jia, M.~Yaghini, C.~A. Choquette-Choo, N.~Dullerud, A.~Thudi,
  V.~Chandrasekaran, and N.~Papernot, ``Proof-of-learning: Definitions and
  practice,'' in \emph{2021 IEEE Symposium on Security and Privacy (SP)}.\hskip
  1em plus 0.5em minus 0.4em\relax IEEE, 2021, pp. 1039--1056.

\bibitem{zhang2022adversarial}
R.~Zhang, J.~Liu, Y.~Ding, Z.~Wang, Q.~Wu, and K.~Ren, ``“adversarial
  examples” for proof-of-learning,'' in \emph{2022 IEEE Symposium on Security
  and Privacy (SP)}.\hskip 1em plus 0.5em minus 0.4em\relax IEEE, 2022, pp.
  1408--1422.

\bibitem{shumailov2021manipulating}
I.~Shumailov, Z.~Shumaylov, D.~Kazhdan, Y.~Zhao, N.~Papernot, M.~A. Erdogdu,
  and R.~J. Anderson, ``Manipulating sgd with data ordering attacks,''
  \emph{Advances in Neural Information Processing Systems}, vol.~34, pp.
  18\,021--18\,032, 2021.

\bibitem{gao2022verifi}
X.~Gao, X.~Ma, J.~Wang, Y.~Sun, B.~Li, S.~Ji, P.~Cheng, and J.~Chen, ``Verifi:
  Towards verifiable federated unlearning,'' \emph{arXiv preprint
  arXiv:2205.12709}, 2022.

\bibitem{sommer2022athena}
D.~M. Sommer, L.~Song, S.~Wagh, and P.~Mittal, ``Athena: Probabilistic
  verification of machine unlearning,'' \emph{Proceedings on Privacy Enhancing
  Technologies}, vol.~3, pp. 268--290, 2022.

\bibitem{weng2022proof}
J.~Weng, S.~Yao, Y.~Du, J.~Huang, J.~Weng, and C.~Wang, ``Proof of unlearning:
  Definitions and instantiation,'' \emph{arXiv preprint arXiv:2210.11334},
  2022.

\bibitem{lundberg2017unified}
S.~M. Lundberg and S.-I. Lee, ``A unified approach to interpreting model
  predictions,'' \emph{Advances in neural information processing systems},
  vol.~30, 2017.

\bibitem{takahashi2017aenet}
N.~Takahashi, M.~Gygli, and L.~Van~Gool, ``Aenet: Learning deep audio features
  for video analysis,'' \emph{IEEE Transactions on Multimedia}, vol.~20, no.~3,
  pp. 513--524, 2017.

\bibitem{ngiam2011multimodal}
J.~Ngiam, A.~Khosla, M.~Kim, J.~Nam, H.~Lee, and A.~Y. Ng, ``Multimodal deep
  learning,'' in \emph{Proceedings of the 28th international conference on
  machine learning (ICML-11)}, 2011, pp. 689--696.

\bibitem{ioffe2015batch}
S.~Ioffe and C.~Szegedy, ``Batch normalization: Accelerating deep network
  training by reducing internal covariate shift,'' in \emph{International
  conference on machine learning}.\hskip 1em plus 0.5em minus 0.4em\relax pmlr,
  2015, pp. 448--456.

\bibitem{gupta2021adaptive}
V.~Gupta, C.~Jung, S.~Neel, A.~Roth, S.~Sharifi-Malvajerdi, and C.~Waites,
  ``Adaptive machine unlearning,'' \emph{Advances in Neural Information
  Processing Systems}, vol.~34, pp. 16\,319--16\,330, 2021.

\bibitem{jagielski2020auditing}
M.~Jagielski, J.~Ullman, and A.~Oprea, ``Auditing differentially private
  machine learning: How private is private sgd?'' \emph{Advances in Neural
  Information Processing Systems}, vol.~33, pp. 22\,205--22\,216, 2020.

\bibitem{jayaraman2019evaluating}
B.~Jayaraman and D.~Evans, ``Evaluating differentially private machine learning
  in practice,'' in \emph{28th USENIX Security Symposium (USENIX Security 19)},
  2019, pp. 1895--1912.

\bibitem{hu2015differential}
X.~Hu, M.~Yuan, J.~Yao, Y.~Deng, L.~Chen, Q.~Yang, H.~Guan, and J.~Zeng,
  ``Differential privacy in telco big data platform,'' \emph{Proceedings of the
  VLDB Endowment}, vol.~8, no.~12, pp. 1692--1703, 2015.

\bibitem{shannon1948mathematical}
C.~E. Shannon, ``A mathematical theory of communication,'' \emph{The Bell
  system technical journal}, vol.~27, no.~3, pp. 379--423, 1948.

\bibitem{eisenhofer2022verifiable}
T.~Eisenhofer, D.~Riepel, V.~Chandrasekaran, E.~Ghosh, O.~Ohrimenko, and
  N.~Papernot, ``Verifiable and provably secure machine unlearning,''
  \emph{arXiv preprint arXiv:2210.09126}, 2022.

\bibitem{thudi2022unrolling}
A.~Thudi, G.~Deza, V.~Chandrasekaran, and N.~Papernot, ``Unrolling sgd:
  Understanding factors influencing machine unlearning,'' in \emph{2022 IEEE
  7th European Symposium on Security and Privacy (EuroS\&P)}.\hskip 1em plus
  0.5em minus 0.4em\relax IEEE, 2022, pp. 303--319.

\bibitem{tishby2000information}
N.~Tishby, F.~C. Pereira, and W.~Bialek, ``The information bottleneck method,''
  \emph{arXiv preprint physics/0004057}, 2000.

\bibitem{molnar2020interpretable}
C.~Molnar, \emph{Interpretable machine learning}.\hskip 1em plus 0.5em minus
  0.4em\relax Lulu. com, 2020.

\end{thebibliography}

\vfill

\end{document}